\documentclass{article}

\PassOptionsToPackage{numbers,sort,compress}{natbib}

\usepackage[preprint]{neurips_2024}

\usepackage{tikz}
\usetikzlibrary{calc}
\usepackage[utf8]{inputenc} 
\usepackage[T1]{fontenc}    
\usepackage{hyperref}       
\usepackage{url}            
\usepackage{booktabs}       
\usepackage{amsfonts}       
\usepackage{nicefrac}       
\usepackage{microtype}      
\usepackage{xcolor}         
\usepackage{soul}
\usepackage{graphicx}
\usepackage{amsmath,amssymb,amsfonts}
\usepackage{bbm}
\usepackage{wrapfig}
\usepackage{enumitem}
\usepackage{tcolorbox}

\definecolor{britishracinggreen}{rgb}{0.0, 0.26, 0.15}

\newcommand{\ex}{\mathbb{E}}

\newcommand{\info}{\mathcal{I}}
\newcommand{\pred}{\hat{\mathbf{p}}}

\newcommand{\ind}{\mathbbm{1}}
\usepackage{mathtools}
\newcommand\model{\stackrel{\mathclap{\normalfont\mbox{\tiny{m}}}}{=}}

\usepackage{amsmath,amsfonts,bm}

\def\eqref#1{equation~\ref{#1}}

\def\1{\bm{1}}

\DeclareMathAlphabet{\mathsfit}{\encodingdefault}{\sfdefault}{m}{sl}
\SetMathAlphabet{\mathsfit}{bold}{\encodingdefault}{\sfdefault}{bx}{n}

\newcommand{\real}{\mathbb{R}}

\DeclareMathOperator*{\argmin}{arg\,min}

\title{Predicting Probabilities of Error to Combine Quantization and Early Exiting: QuEE}

\author{%
 Florence Regol\\
 McGill University, ILLS\thanks{International Laboratory on Learning Systems}, MILA\thanks{Quebec Institute for Learning Algorithms}\\
 \texttt{florence.robert-regol@mail.mcgill.ca}
 \And
 Joud Chataoui\\
 McGill University, ILLS, MILA\\
 \texttt{joud.chataoui@mail.mcgill.ca}
 \And
  Bertrand Charpentier\\
  Technical University of Munich\\
 \texttt{charpent@in.tum.de}
 \And
 Mark Coates\\
 McGill University, ILLS, MILA\\
 \texttt{mark.coates@mcgill.ca}
 \And
 Pablo Piantanida\\
 École de technologie supérieure, ILLS, MILA\\
 \texttt{pablo.piantanida@mila.quebec}
 \And
 Stephan Gunnemann\\
 Technical University of Munich\\
 \texttt{guennemann@in.tum.de}
}

\begin{document}

\maketitle

\begin{abstract}
Machine learning models can solve complex tasks but often require significant computational resources during inference. This has led to the development of various post-training computation reduction methods that tackle this issue in different ways, such as quantization which reduces the precision of weights and arithmetic operations, and dynamic networks which adapt computation to the sample at hand. In this work, we propose a more general dynamic network that can combine both quantization and early exit dynamic network: QuEE. Our algorithm can be seen as a form of soft early exiting or input-dependent compression. Rather than a binary decision between exiting or continuing, we introduce the possibility of continuing with reduced computation. This complicates the traditionally considered early exiting problem,  which we solve through a principled formulation. The crucial factor of our approach is accurate prediction of the potential accuracy improvement achievable through further computation.
We demonstrate the effectiveness of our method through empirical evaluation, as well as exploring the conditions for its success on 4 classification datasets.
\end{abstract}

\section{Introduction}

Large models, paired with transfer learning or fine-tuning, are becoming established as a dominant approach in machine learning \citep{gpt3, devlin2019bert, oquab2024dinov2, dettmers2022_llmint8}. Consequently, reducing the inference cost of large pretrained models without (or with minimal) retraining is becoming increasingly important~\citep{dettmers2022_llmint8, loftq_lora, efficient_ptq_lm}.

\begin{figure}[ht]
    \centering
    \includegraphics[scale=0.58]{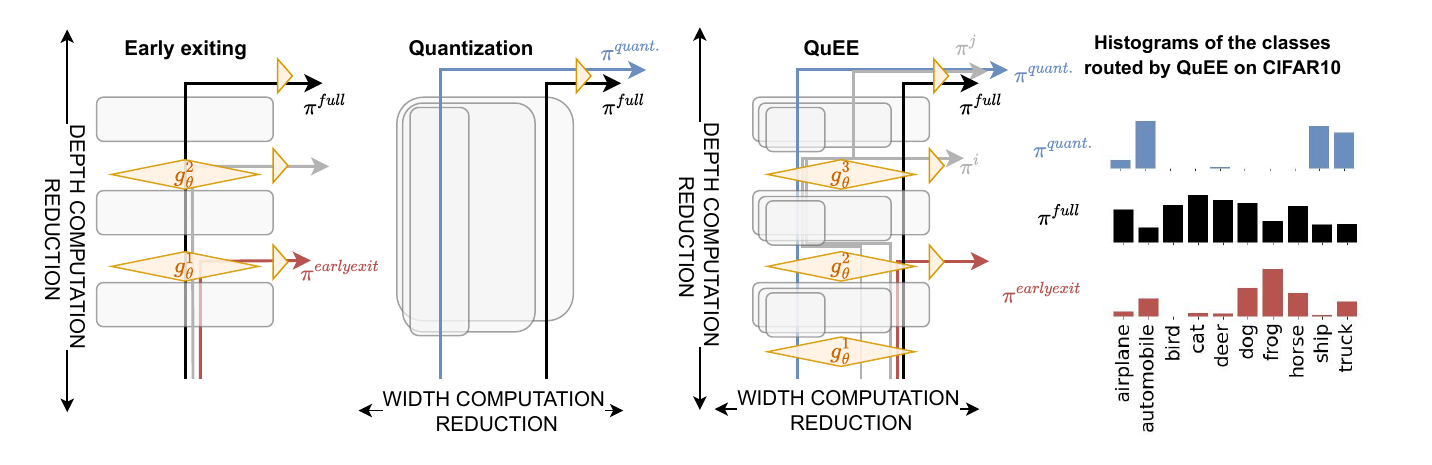}
    \caption{ Early exit decreases inference costs along the depth, while quantization decreases inference costs along the width. QuEE can integrate both, and learn to employ different computation reduction methods to classify different classes. On CIFAR-10, quantization is mainly used for automobiles and ships, while early exiting is mainly used for dogs and frogs.  }
    \label{fig:arch}
\end{figure}
There is a wide array of approaches to post-training computation reduction: quantization reduces the precision of stored weights and activations \citep{compression_survey, PTQ4ViT, efficient_ptq_lm}, distillation trains a smaller model to imitate a larger one \citep{hinton2015distilling, knowledge_distillation_survey}, pruning removes weights or units entirely \citep{compression_survey}, and dynamic networks  can adapt the computation to the sample at hand~\citep{han2022_dynamic_survey}.
These computation reduction methods vary significantly and offer different benefits. Our aim in this work is to develop a method that can combine different approaches to leverage their complementary strengths.



Although there have been proposals to combine approaches, including the application of fixed quantization to early exit networks~\citep{saxena2023_mcqueen,escepe} and sample-aware quantization~\citep{Liu2022_dqnet, hong2022cadyq, Tian2023_cabm}, there have been few efforts towards selecting and combining, on a per-sample basis, multiple post-training cost reduction techniques. Our goal is to design a unified framework that determines which combination of computation reduction methods should be employed for each sample, in a fully post-training setting. In our proposed dynamic network, computation reduction can be performed in two ways: both in depth and in width (see Figure~\ref{fig:arch}). We achieve this by combining early exiting for depth adaptation with quantization for width adaptation. We opt for quantization as it currently offers more practical and superior compression results post-training without fine-tuning~\citep{kuzmin2023pruning}. In contrast, pruning and distillation have received more attention in the fine-tuning setting~\citep{Wang2024StructurallyPA,frantar2023gpt}.


This allowance for adaptive per-sample quantization/early exiting significantly complicates the problem. Traditionally, in early exiting, the decision that must be made for each sample at each candidate exit point is whether to exit or continue. This permits the application of a simple binary threshold-based rule \cite{msdnet, dynamic_perc, bolukbasi_adaptive_nn}. Alternatively, we can train exit-controller modules \cite{regol2024_jeidnn, epnet_rl_learnable, Ilhan2023AdaptiveDN}.
Such an approach is infeasible in the new setting due to the increased number of options (exit or quantize at multiple different levels). We address the problem with a principled approach to solve the more general dynamic network problem, drawing on insights from theoretical works~\cite{verma2023, jitkrittum2023when}.

In particular, we observe that solving the dynamic network problem for fixed classifiers can be addressed by solving the task of predicting the probabilities of error of the candidate downstream (more computationally expensive) classifiers. 
Experimentally, we demonstrate that the proposed procedure can generate sufficiently accurate estimates of the probabilities of error, allowing the per-sample selection of appropriate levels of quantization and effective exit points. 

Our paper makes the following contributions:
\begin{enumerate}[leftmargin=*]
    \item We introduce a novel and principled view of the dynamic network problem formulation,  and reframe it as a task of predicting error probabilities.
    \item To learn to predict  the inaccessible error probabilities, we introduce a new method that involves discretizing the feature space and using empirical error  approximations as our training targets.
    \item Our unifying formulation allows us to to combine different computation reduction methods in a post training setting, leading to the introduction of a new learning framework: QuEE. 
    \item We empirically demonstrate the efficacy of our method, and we explore necessary conditions for its effectiveness, using 4 classification datasets.
\end{enumerate}


\section{Related Work} 
\paragraph{Dynamic neural networks.} Dynamic architectures adapt their computational graphs to the input\cite{han2022_dynamic_survey, compression_survey}. By adapting depth (a subset of layers are executed) or width (a subset of channels or neurons executed) for each sample, dynamic architectures can reduce computation during inference \cite{han2022_dynamic_survey}.

Early Exit (EE) networks are a class of dynamic depth architecture where a prediction is obtained at an intermediate layer and subsequent layers are skipped \cite{bolukbasi_adaptive_nn, msdnet, dynamic_perc, regol2024_jeidnn, Ilhan2023AdaptiveDN}. This is done by augmenting the network with intermediate inference modules at various layers. Early works propose architectures tailored for EE that are trained end-to-end, often paired with a simple thresholding mechanism as an exit rule~\citep{bolukbasi_adaptive_nn,msdnet,dynamic_perc}. This approach can provide significant efficiency savings, but end-to-end training can be impractical for large models~\cite{efficient_ptq_lm, regol2024_jeidnn}.
To address this, post-training EE methods that rely on a fixed pre-trained backbone have been introduced -- they instead explore effective ways to train the inference modules and design more sophisticated gating mechanisms for the exit rule~\citep{regol2024_jeidnn,epnet_rl_learnable, Ilhan2023AdaptiveDN}.  
However, these often come with the drawback of having to repeat the training process for every operating point, making it impractical for use-cases where the computation budget changes over time.

Width-wise sample-adaptation can be achieved by selecting a subset of channels in CNNs~ \citep{huang2018condensenet, herrmann2020_channel_selection_gumbel}. A more general approach, compatible with transformer-based architectures, is SuperNets \cite{liu2018_d2nn, odena2017_changing_model_behavior_rl, hazimeh2020_tree_ensemble_layer} where samples are dynamically routed through a subset of neurons at inference. 
Closer to our algorithm, works such as \citet{dual_dynamic_inference, fully_dynamic_inference_dnn} perform both depth and width adaptation via layer-skipping and channel-selection in convolution-based architectures. These approaches rely on trainable controllers that are optimized jointly with the underlying network \cite{dual_dynamic_inference, fully_dynamic_inference_dnn}. This makes them incompatible with large pre-trained models~\cite{vit_survey, touvron2023_llama2}.

\paragraph{Quantization.} Quantization is another effective way of speeding up inference. Weights, gradients and activations of a model are represented at lower bit resolutions \cite{compression_survey}. Quantization-aware training (QAT) techniques quantize the network during training \cite{quant_survey, quant_survey_gholami}, while post-training quantization (PTQ) is performed on a trained model~\cite{quant_survey, quant_survey_gholami,dettmers2022_llmint8}. 
This makes PTQ particularly appealing as a width-compression technique for our setting, i.e., working with large pre-trained models \cite{efficient_ptq_lm, dettmers2022_llmint8}.
However, quantization is typically not input-adaptive~\cite{quant_survey, Liu2022_dqnet, hong2022cadyq}. There are a few exceptions~\cite{hong2022cadyq, Tian2023_cabm, Liu2022_dqnet}. ~\citet{hong2022cadyq} and ~\citet{Tian2023_cabm} consider dynamic mixed-precision quantization for the specific problem of image super-resolution.  Better suited for our more general setting, DQNet~\cite{Liu2022_dqnet} explores dynamic mixed-precision quantization for image classification. In DQNet, the network is augmented with a small neural network called the \textit{bit controller} whose role is to determine the bit resolution of each layer for a given sample. 
While \citet{Liu2022_dqnet, hong2022cadyq}, and  \citet{Tian2023_cabm} all dynamically adapt bit precision on a per-sample basis, they encode the computational budget in their loss formulation, meaning that the algorithm needs to be retrained for every operating point.

\paragraph{Quantization of early exit networks.} Several works combine the adaptability of early exit networks with the efficacy of quantization by quantizing early exit networks \cite{saxena2023_mcqueen, escepe}. In \citep{escepe}, a pre-trained early exit network is first split into sections. Each section is then quantized separately using weight-clustering. The quantized network is fully retrained using knowledge distillation. \citet{saxena2023_mcqueen} use a QAT approach, where the optimal per-layer quantization parameters are learnt during training. While these works combine quantization with early exiting, they both propose QAT-like approaches, and are thus unsuitable for large models. They are also not sample-adaptive along their widths, as a single static mixed-precision quantization is learnt for all samples \cite{escepe, saxena2023_mcqueen}.

\section{Problem Setting - Dynamic Networks}

We consider a dynamic network setting with fixed classifiers. Suppose we have access to $M$ computational units: $v_1,\dots,v_M$. These  units can be composed to form $L$ classifiers, $f_1, \dots, f_L$, where $f: \mathcal{X} \to \mathcal{Y}$. For example, we may have $f_1 = v_{M-1}\circ v_2 \circ v_1$ and $f_2 = v_{M} \circ v_2 \circ v_1$. The classifiers have associated costs of evaluation $(c_1, \dots, c_L)$. The cost for classifier $f_l$ is equal to the sum of the costs of its constituent computational units; we assume these costs to be unaffected by $x$. 

To select which classifier will perform the inference for a given sample $x$, we employ a classifier-selector function $S(x) : \mathcal{X} \to [L]$. The classifier-selector function has a sample-dependent cost of evaluation, but it is negligible compared to the cost of the classifiers. We note that $S(x)$ can be the result of a sequence of decisions, interspersed with computation\footnote{as is the case in an early exit dynamic network setting.}. For example, $S(x)$ may first choose to apply computational unit, $v_1$, and then make a second decision, deciding whether to apply computational unit $v_2$ or $v_3$. The results of intermediate computation  can be used when making the later decisions.

The goal is to learn the classifier-selector function $S$ that gives the best performance/computation cost trade-off, which can be quantified in various ways. In this work, we consider a cost-based 01 loss:
\begin{align}
\ell_{01c}(x,y, S) =  \ind[ f_{S(x)}(x) \neq y] +  c_{S(x)}.
\end{align} 
This loss is attractive because it is interpretable; we assign a cost of 1 to classification error and a cost of 0 to correct classification, and then $c_{S(x)}$ controls the penalty associated with computation using the classifiers selected by $S(x)$. 

Our objective is then to parameterize $S$ and find the parameters that minimize this loss in expectation:
\begin{align}
    \theta^* =& \argmin_{\theta} \ex_{XY}\left[ \ind[ f_{S_{\theta}(X)}(X) \neq Y] + c_{S_{\theta}(X)} \right].
\end{align}
We can show from a straightforward extension of results in ~\cite{jitkrittum2023when} and~\cite{verma2023} that the optimal classifier-selector function $S^*(x)$ selects the classifier $f_l$ that has the smallest sum of evaluation cost $c_l$ and probability of making an error, which we denote  by $PE(f_l(x)|x)$:
\begin{align}
    S^*(x) =& \argmin_{S_{\theta}(x)} \ex_{Y|x} \left[ \ind[ f_{S_{\theta}(x)}(x) \neq Y] + c_{S_{\theta}(x)} \right]\\
   =&  \argmin_{l \in [L]} \, \big\{ c_l + 1- \Pr(Y=f_l(x)|x)\big\} \\
     S^*(x)  =&  \argmin_{l \in [L]} \, \{ c_l + PE(f_l(x)|x) \}. \label{eq:optimS}
\end{align}
This solution provides insight into what the classifier-selector function $S$ should achieve:

\begin{tcolorbox}[colback=white!5!white,colframe=red!75!black]
Since the $c_l$ are known, solving the dynamic network problem with fixed classifiers and associated $\ell_{01c}(x, y, S)$ loss can be addressed by accurately predicting $\{PE(f_l(x)|x)\} \, \forall l \in [L] $ -- the probability of making an error for each classifier.\footnotemark  
\end{tcolorbox}
\footnotetext{For each pair of classifiers, $f_{l_1}$ and $f_{l_2}$, with $c_{l_2}> c_{l_1}$, we must determine for each sample whether the  differential of error probabilities: $PE(f_{l_1}(x)|x)-PE(f_{l_2}(x)|x)$ exceeds that of computational costs:  $c_{l_2}-c_{l_1}$.}

\section{Methodology - QuEE}
Now that we have introduced the general framework for dynamic networks with fixed classifiers, we specify how each of the components, $f_l$, $c_l$, and $S_{\theta}(x), \info$, are defined for our proposed architecture, \textbf{QuEE}, in which we mix different computation levels (through quantization) with early exiting. 

\paragraph{Classifiers and costs $(f_l, c_l)$.}
In our proposed dynamic network architecture, we can control both the number of blocks (a block consists of one or more network layers) that are evaluated (using early exiting) and the amount of computation conducted within each block. Each classifier is therefore defined by a ``path'' $\pi$ that traverses blocks of the network, where the number of steps in the path corresponds to the number of blocks evaluated. Denoting by $b_i$ the computation level at block $i$, we can express a path that traverses $e$ blocks before exiting as:
\begin{align}
    \pi &=  b_1 \rightarrow \dots \rightarrow b_{e}.
\end{align}
We denote by $\pi[:j]$ the first $j$ steps of this path ($b_1 \rightarrow \dots \rightarrow b_j$). Given a possible number of exits $E$ and per-block computation levels $B$, the number of paths we can take in the network is $\sum^E_{e=1}B^{e}$. This corresponds to the total number of different classifiers that we could use for a given sample. We denote the set of all possible paths by $\mathcal{P}$. 

We introduce the notation $\pred_{\pi i} \in [0,1]^{|\mathcal{Y}|}$ for the predicted probability vector of classifier $f_{\pi}(x_i)$. If it is not necessary, the index of the input $x_i$ is omitted and we only write $\pred_{\pi}$.

\paragraph{Available pre-computed quantities $\info$.}  At each candidate exit point, a gate $g_{\theta}^j()$ is used to make the decision about how to process the sample.  Before obtaining the decision of gate $g_{\theta}^j()$ for $j>1$, we evaluate the classifier $f_{\pi[:j{-}1]}(x_i)$ that would be used if the sample were to exit. We also place a gate $g_{\theta}^1()$ at the very beginning of the network that can select the computation level of the first block of layers, but that cannot exit ($g_{\theta}^1() \in \{b_1, \dots, b_B\}$).

Therefore, the information $\info_j(x_i)$ available to use as input for each gate $g^j_{\theta}(\cdot )$ includes all the predicted probabilities that were evaluated at previous classifiers and the path followed so far ($\pi[:j{-}1]$):
\begin{align}
\info_j(x_i)  = \begin{cases}
    \{ \pred_{\pi_i[:1] }, \pred_{\pi_i[:2] } , \dots \pred_{\pi_i[:j-1] } ,  \pi_i[:j - 1] \},\quad j > 1\\
    \varnothing,\quad j = 1 \quad \quad \text{(no processed information is available to the first gate)}
\end{cases}
\end{align}

\paragraph{Classifier-selector function  $S_{\theta}(x)$.} 

 In our  setting, the classifier-selector function $S()$ is decomposed into $E{-}1$ gating functions, $g^j_{\theta}()$, for $j=1,\dots,E-1$.  The $j$-th gate can decide to either 1) exit at block $j$, or 2) choose one of the $B$ levels of computation for the next block, i.e., $g^j_{\theta}(\cdot ) : \mathcal{F}_j \to \{0,\dots, b_B\}$, where $0$ indicates exit. We do not specify yet the input space $\mathcal{F}_j$ of the gating functions as it relates to the available information -- we present it in the next section.

The gate $g^j_{\theta}()$ decides the action taken at the $j$-th step:
\begin{align}
   b_{j+1} = \begin{cases} g^j_{\theta}(\cdot ) &\text{ if } g^j_{\theta}(\cdot ) \in \{b_1,\dots, b_B \} \\
   None &\text{ if } g^j_{\theta}(\cdot ) =0 \text{ (exit) } \end{cases}
\end{align}
The path $\pi_i$ selected for a given sample $x_i$ is then determined by this sequence of decisions, with termination when one of the gates decides to exit:
$\pi_i =  g^1_{\theta}(\cdot ) \to  \dots \to g^e_{\theta}(\cdot )=0$, $g^j_{\theta}(\cdot ) \neq 0 \,\,\,\forall \,\,j<e.$
The selected path is then iteratively constructed as we successively evaluate  $g_{\theta}(\cdot )$. 

\paragraph{QuEE  $S_{\theta}(x)$.} 
\begin{wrapfigure}{r}{0.5\linewidth}
   \label{figures/arch/metho.pdf}
   \vspace{-10mm}
    \centering
   \includegraphics[width=0.5\textwidth]{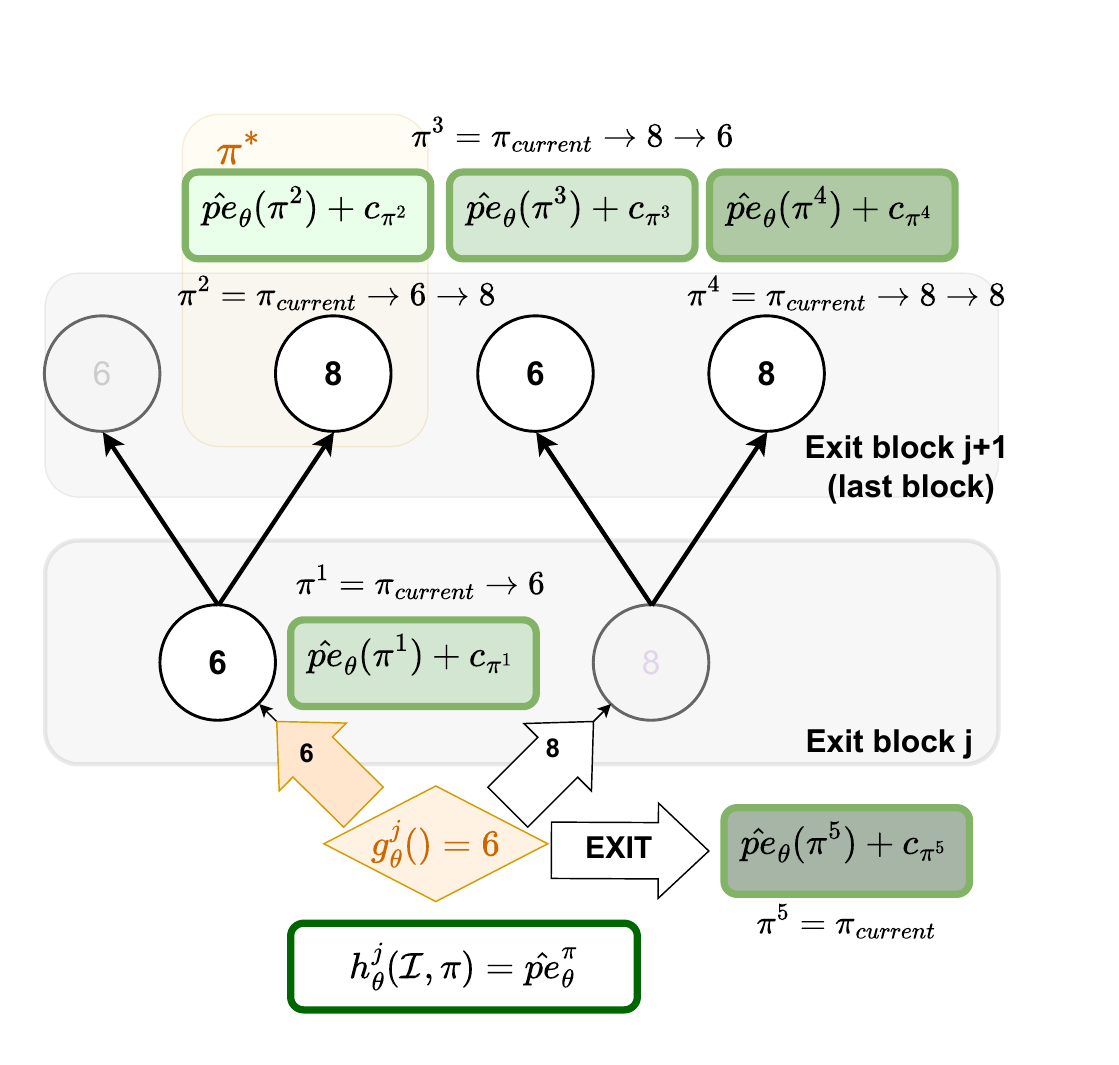}
    \caption{Depiction of how QuEE predicts the next step at inference. $h_{\theta}$ is evaluated for each considered paths (2 paths shown in grey were not sampled), then the algorithm takes a step towards the path $\pi^*$  minimizing the predicted loss $\ell_{01c}$. In this example, at gate $j$  $\pi^2$ has been identified as the best path.}\vspace{-5em}
 \end{wrapfigure}
As presented in the previous section, a successful classifier-selector will accurately predict the probability of future classifiers $f_{\pi}$ being incorrect ($PE(f_{\pi}|x)$) and select the classifier with the lowest combined cost and probability of error (see Eqn~\ref{eq:optimS}). 


Therefore, we propose to explicitly learn to predict the probability of error $\widehat{pe}_{\theta}(f_{\pi'}|x) \model PE(f_{\pi'}|x) $ of each potential path $\pi' \in \mathcal{P}_{\pi[:j - 1]}$ at each decision gate $g^j_{\theta}(\info_j)$.  Each gate is therefore equipped with a predictor module $h^j_{\theta}(\info_j(x), \pi') = \widehat{pe}_{\theta}(f_{\pi'}|x)  $ that takes as input $\info_j$ and the future path to predict $\pi'$. We can then select the step that would lead to the predicted optimal path, i.e., $g^j_{\theta}(x) =\hat{\pi}^*[j{-}1:j]$, where 
\begin{align}
    \hat{\pi}^*(x) &= \argmin_{\pi' \in \mathcal{P}_{\pi[:j-1]} }\,  \big\{ c_{\pi'}+\widehat{pe}_{\theta}(f_{\pi'}|x)\big\}, \nonumber \\\quad \text{ for }&\,\, \widehat{pe}_{\theta}(f_{\pi'}|x) = h^j_{\theta}(\info_j(x), \pi')\,.
\label{eq:prop1_gate}
\end{align}
This has a complexity of $O(EB^E)$ because we must infer for each potential path $\pi' \in \mathcal{P}_{\pi} $. If this is prohibitive, we can sample a subset of paths to evaluate, with $\pi' \sim U[\mathcal{P}_{\pi}] $. An alternative approach is to directly predict the next step to avoid the exponential complexity in inference, but this also has some drawbacks. We explore the trade-off in Appendix~\ref{app:next_best}. A notable advantage of our approach is the complete decoupling of the learning procedure from the assignment of costs to classifiers. The costs $c_{\pi'}$ are determined at inference and can be modified without necessitating any retraining or adjustments.

\subsection{Approximating the probability of error $PE(f_{\pi}|x)$}\label{sec:approximating_prob_error}
Our methodology involves learning to predict the performance we could achieve given more computational power. This would motivate using $PE(f_{\pi}|x)$ as the target values during training, but we do not have access to these quantities. Therefore, we must rely on approximations to set the learning targets for $\widehat{pe}_{\theta}(f_{\pi}|x)$.  There is existing work on estimating these quantities~\cite{granese2021doctor,gomes2024a}, but the proposed procedures tend to be only reliable for high-performing classifiers (we verified this through experiment). The intermediate classifiers at lower blocks are often relatively inaccurate.

We base our approximations on the assumption that $PE(f_{\pi}|x_i)$ is smooth over some transformation of the space $t_{\pi}(x)$. This allows us to discretize the space into  $K$ partitions $\{Q_1, \dots, Q_K\}$ for which, under our assumption, the resulting distances $|PE(f_{\pi}|x_i) - \frac{1}{p_Q}\int_{x \in Q} PE(f_{\pi}|x)P(dx)  |$ are small for all $x_i \in Q$. Here we have introduced $p_Q \triangleq \int_{Q}P(dx)$. 

We can then compute the empirical estimator of $\frac{1}{p_Q}\int_{Q} PE(f_{\pi}|x)P(dx)$ for each partition using a validation set. Assume that we have $m$ samples $\{x_j, y_j\}_{j=1}^m$ in the validation set and denote by $m_Q$ the number of samples in $Q$, i.e. $m_Q = \sum_{j=1}^m \ind[x_j \in Q]$. Then:
\begin{align}
PE(f_{\pi}|x_i) &\approx \frac{1}{\Pr(Q)}\int_{Q} PE(f_{\pi}|x)P(dx), \quad \mathrm{for} \,\, x_i \in Q\,, \\
&\approx \frac{1}{m_Q}\sum_{j=1}^m   \ind[f_{\pi}(x_j) \neq y_j, x_j \in Q]\label{eq:cluster_approx}, \\
  &\triangleq   \widetilde{pe}(f_{\pi}  |x_i)  \text{ our approximation used to train } \widehat{pe}_{\theta}(f_{\pi}|x). \label{eq:approx}
\end{align}

In practice, we discretize the space using a clustering algorithm and we use the predicted probability vectors as the clustering features ($t_{\pi}(x) = \pred_{\pi} $). We provide empirical results and qualitative analysis to motivate this choice in Appendix~\ref{app:cluster} and in the result section.

\subsection{Modeling the probability of error with $h^j_{\theta}(\info_j(x), \pi')$}

The learning task of $h^j_{\theta}(\info_j(x), \pi')$ is to predict the ground truth approximation $\widetilde{pe}(f_{\pi}  |x)$ of future classifiers $f_{\pi'}$. Each classifier is encoded by its associated path $\pi'$, and we have access to past predicted probability vectors $\{ \pred_{\pi_i[:1] }, \pred_{\pi_i[:2] } , \dots \pred_{\pi_i[:j] }  \}$  through $\info_j(x_i)$.
We cannot employ a complex architecture for $h^j_{\theta}(\info_j(x), \pi')$ because the goal is to reduce computational overhead.

We base the decision on the current probability vector $\pred_{\pi_i[:j] }$ and the previous one $\pred_{\pi_i[:j-1]}$. We additionally extract helpful statistics such as entropy $H(\pred_{\pi_i[:j] }), H(\pred_{\pi_i[:j-1] })$ and maximum probability, $\max (\pred_{\pi_i[:j] }), \max(\pred_{\pi_i[:j-1]})$. These are often used in early exiting algorithms \cite{bolukbasi_adaptive_nn, dynamic_perc, msdnet, boostedNet}. The final vector $u_i \in \real^{2|\mathcal{Y}| + 4}$ that encodes all the input features is:
\begin{align}
    u_i &= \pred_{\pi_i[:j] }\,||\, \pred_{\pi_i[:j-1]} \,||\, H(\pred_{\pi_i[:j] }) \,||\, H(\pred_{\pi_i[:j-1] })\,||\, \max (\pred_{\pi_i[:j-1] }) \,||\, \max (\pred_{\pi_i[:j] })\,.
\end{align}
As for the path  $\pi'$, we encode it in a fixed length vector $p_i$ of size $E{+}1$ by padding shorter paths with 0s, and add the number of layers evaluated at the end:
\begin{align}
    p_i &= [b_1, b_2, \dots, b_e, 0, 0,e] \in \real^{E+1}.
\end{align}
Our decision architecture is a simple 2-layer network:
\begin{align}
\widehat{pe}_{\theta}(f_{\pi'}|x) =  \mathrm{sigmoid}(NN(NN(u_i || p_i)))\label{eqn:arch_h},
\end{align} 
and we train each predictor module $h^j_{\theta}(\info_j(x), \pi')$ with a standard MSE loss:
\begin{align}
    \mathcal{L}(x, \pi') =  || h^j_{\theta}(\info_j(x), \pi') -  \widetilde{pe}(f_{\pi'}  |x))||_2. \label{eq:loss}
\end{align}

\section{Experiments}

We present results on image classification tasks: ImageNet~\citep{imagenet}, CIFAR10 and CIFAR100~\citep{krizhevsky_cifar}, and SVHN~\citep{svhn}, for different backbone architectures: T2T-ViT~\citep{yuan2021_t2t_vit} pre-trained on ImageNet and transfer-learnt on the datasets, and ViT-14 \cite{vit}, pre-trained as a foundation model using DinoV2~\cite{oquab2024dinov2}.

\subsection{Obtaining the computation units for the classifiers $f_{\pi}$}

We start by augmenting the pre-trained backbone with randomly initialized classifiers at predefined layers (See App. \ref{app:exp_details} for details about exit placement). The classifiers have the same architecture as the final inference head. We train intermediate classifiers until convergence, keeping the backbone frozen. We then quantize the network at various bit-widths corresponding to our choices of computation levels, $\{b_1,\dots, b_B\}$, as follows. The highest bit-width in $b_B$ is chosen to be the lowest value that maintains the accuracy of the final classifier (usually around 8 bits). We then add decreasing bit-widths to $b_B$ as long as the accuracy of the final classifier is within 5\% of the original performance. Typically, we obtained at most $B = 4$. We perform the quantization at each bit-width using PTQ4ViT~\cite{PTQ4ViT}, a state-of-art PTQ algorithm for transformers (See Appendix \ref{app:baselines} for an in-depth description). The network obtained is thus a multi-quantization architecture comprised of L blocks augmented with an inference head, where each block consists of the backbone layer quantized at the B levels.

\textbf{QuEE components} We finally augment the multi-quantization network with gates $\{g_{\theta}^j\}_{j \in E}$ at every exit. Each gate $g_{\theta}^j$ contains a small learnable predictor module $h_{\theta}^j$. The first gate  $g_{\theta}^1$ is placed at the start of the network and always routes to the highest computation level.

\subsection{Training the predictor modules $h_{\theta}^j$}
Before training the predictor modules, we fit a K-means algorithm for every path in the network in order to obtain discretized predictor targets, where $K$ is a hyperparameter. In practice, we observe that paths that move from a low bit resolution to a higher bit resolution result in subpar performance. We thus do not consider these paths during training and inference. This effectively reduces the set of paths $\mathcal{P}$ by a factor of two.  If the resulting $|\mathcal{P}|>50$, we randomly sample 50 paths, both during training and at inference. For each valid path $\pi$, we obtain $K$ clusters by clustering the predicted probability vectors $\pred_{\pi} \in [0,1]^{|\mathcal{Y}|}$ using  the validation set. For each cluster $Q$ we compute $\widetilde{pe}(f_{\pi, Q}  |x_i)$ using  \eqref{eq:cluster_approx}. At training, we obtain the discretized targets by predicting the appropriate cluster $Q^i$ for $\pred_{\pi'}$ and using $\widetilde{pe}(f_{\pi, Q^i}  |x_i)$ as the target in the MSE loss~(\eqref{eq:loss}). This is optimized with the Adam optimizer and early stopping (using the validation set). The complete set of optimization hyperparameters is provided in Appendix~\ref{app:exp_details}.

\subsection{Routing at inference}\label{sec:exp:routing_at_inf}
At inference,  when the sample $x$ reaches $g^2$, the probability of future error $\widehat{pe}_{\theta}(f_{\pi'}|x)$ for every path $\pi' \in \mathcal{P}_{\pi[1]}$ is computed and $x$ is forwarded along the optimal path, as stated in equation \ref{eq:prop1_gate}. The cost $c_{\pi'}$ of each path in equation \ref{eq:prop1_gate} is the normalized cost in BitOPS \cite{Liu2022_dqnet, Tian2023_cabm, adabits}, computed as follows. For each layer $l$ in $\pi' = b_1 \rightarrow  \dots \rightarrow b_e$, we compute the layer cost $c_l = \text{BitOPS}(l) = \text{FLOPS}(l) \times b_l$. The unnormalized cost of the path is $c_{\pi', \text{unnorm}} = \sum_{l = 1}^e c_{l}$. We normalize all path costs by dividing them by the cost of the most costly path, $\pi_{\text{max}} = b_{\text{max}} \rightarrow b_{\text{max}}\dots$, which is incurred when evaluating the entire network using the highest bit-width $b_{\text{max}}$. Therefore:
\begin{equation}\label{eq:normalized_cost}
    c_{\pi} =\frac{c_{\pi', \text{unnorm}}}{c_{\pi_{\text{max}}}} = \frac{\sum_{l = 1}^e \text{FLOPS}(l) \times b_l}{\sum_{l = 1}^L \text{FLOPS}(l) \times b_{\text{max}}}\,.
\end{equation}
At inference, we can obtain different cost-accuracy operating points without retraining QuEE by incorporating a cost-importance hyper-parameter $\lambda \in \mathbb{R}^+$ in equation \ref{eq:prop1_gate}. This allows us to prioritize efficiency or accuracy:
\begin{equation}
    \hat{\pi}_{\text{adaptive}}^*(x) = \argmin_{\pi' \in \mathcal{P}_{\pi[:j-1]} }\,  \big\{ c_{\pi'} \lambda+\widehat{pe}_{\theta}(f_{\pi'}|x)\big\}\,.
\end{equation}
\vspace{-2em}
\subsection{Baselines}
As our proposal combines early exiting and quantization, we include baselines from both settings.

\begin{itemize}[leftmargin=*]
    \item \textbf{DQNet-gate}~\cite{Liu2022_dqnet} is a data-adaptive mixed-precision quantization architecture. It is a modified version of DQNet~\cite{Liu2022_dqnet}, where the parameters of the backbone are fixed instead of being trained alongside the bit-selector module. We achieve different accuracy/cost points by changing the target bit-width and cost importance parameter $\alpha$.
    
\item \textbf{PTQ4ViT}~\cite{PTQ4ViT} is a SOTA PTQ method for ViT models. We report the accuracy obtained a various levels of quantization. 

\item \textbf{JEIDNN}\cite{regol2024_jeidnn} is a SOTA EE method for frozen backbones. As a backbone model, we use the backbone quantized at the highest precision bit-width we consider.

\item \textbf{Thresholding} applies the most popular gating decision used in the EE literature, where we exit if the max probability exceeds a certain threshold. 
\end{itemize}

\subsection{Results}
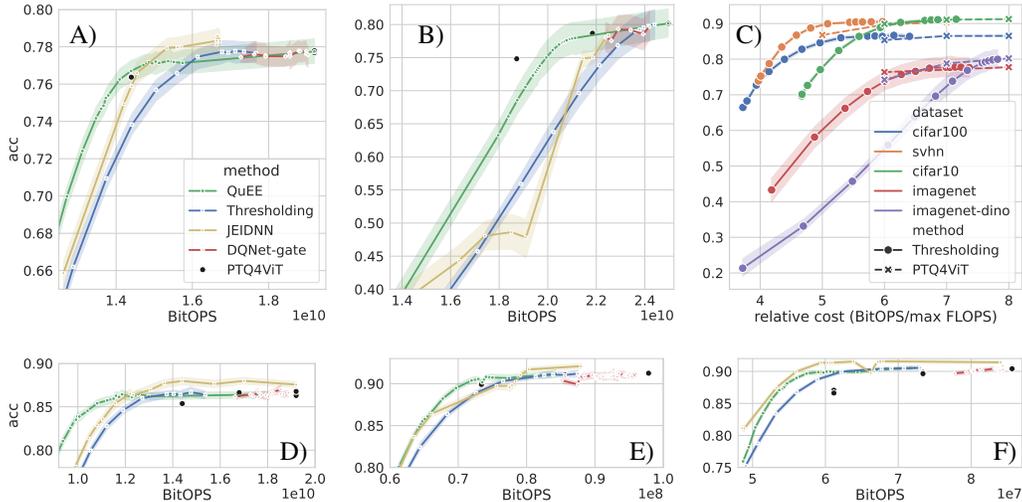
\begin{figure}[h]
    \centering
    \begin{minipage}[t]{0.325\linewidth}
     \begin{tikzpicture}
  \node[inner sep=0pt] (A) { \includegraphics[scale=0.23]{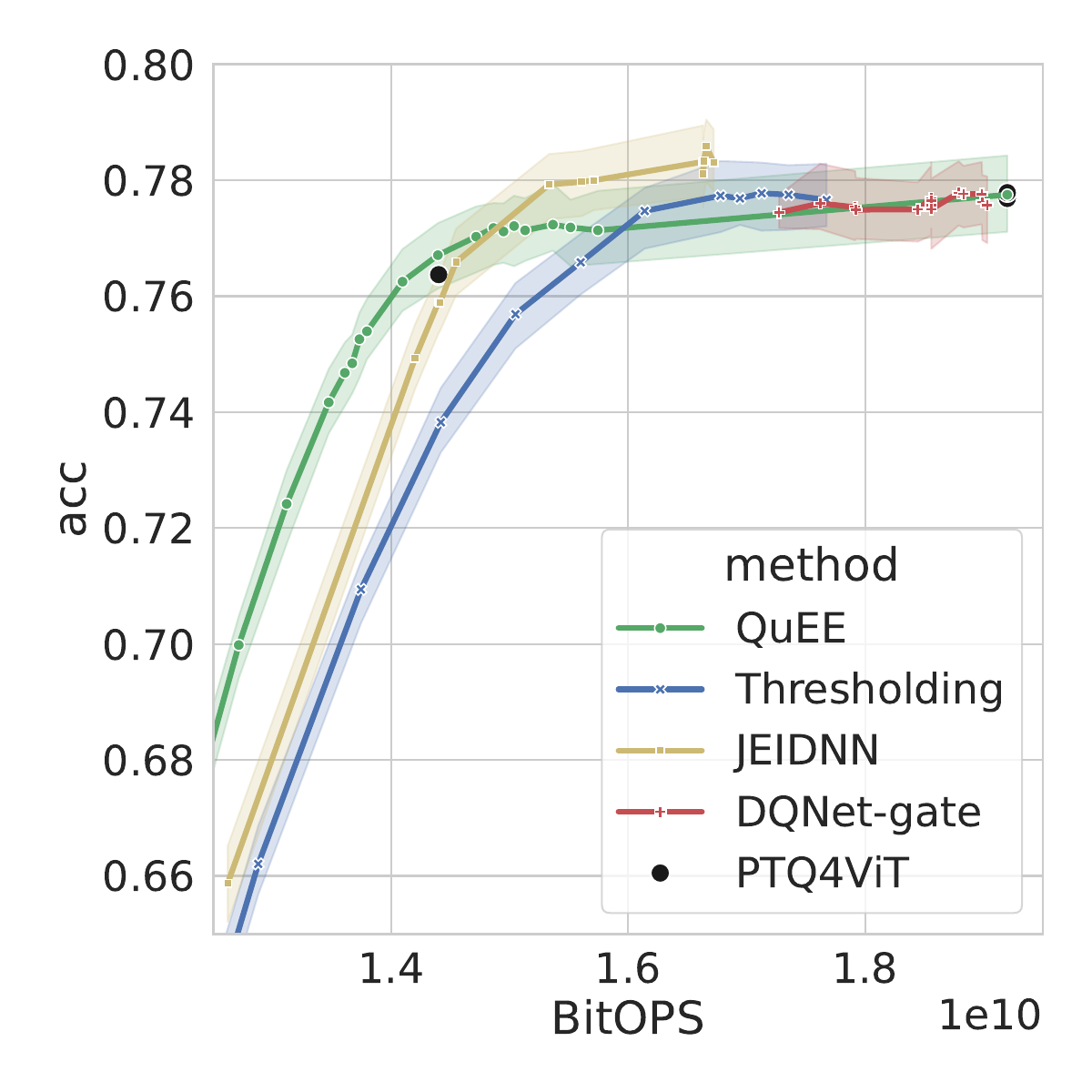}};
  \node[black] at (-1.1,1.7) {A)};
\end{tikzpicture}
    \end{minipage}
    \begin{minipage}[t]{0.325\linewidth}
         \centering
          \begin{tikzpicture}
  \node[inner sep=0pt] (A) { \includegraphics[scale=0.23]{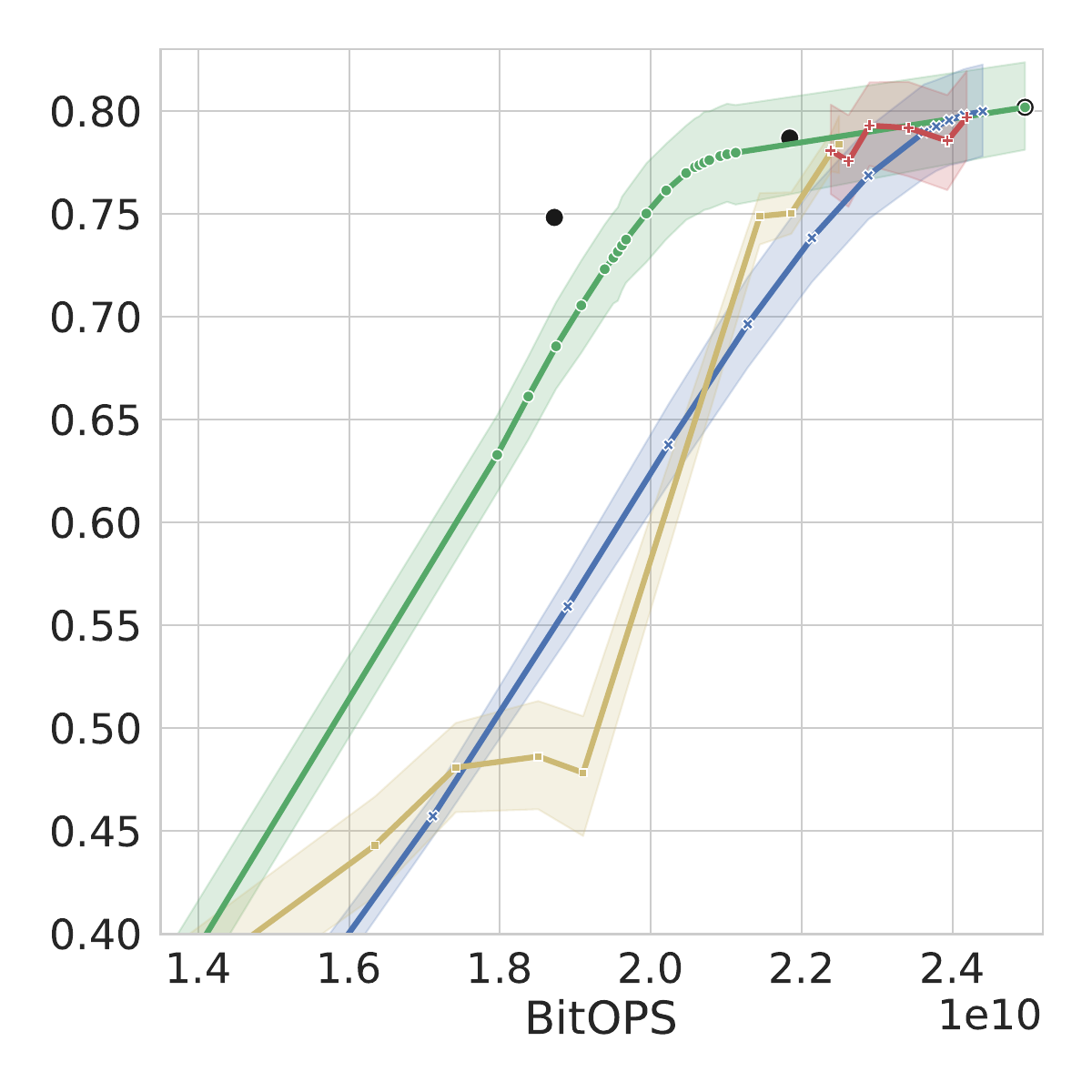}};
  \node[black] at (-1.1,1.7) {B)};
\end{tikzpicture}
    \end{minipage}\label{fig:result_top}
      \begin{minipage}[t]{0.325\linewidth}
        \begin{tikzpicture}
  \node[inner sep=0pt] (A) { \includegraphics[scale=0.23]{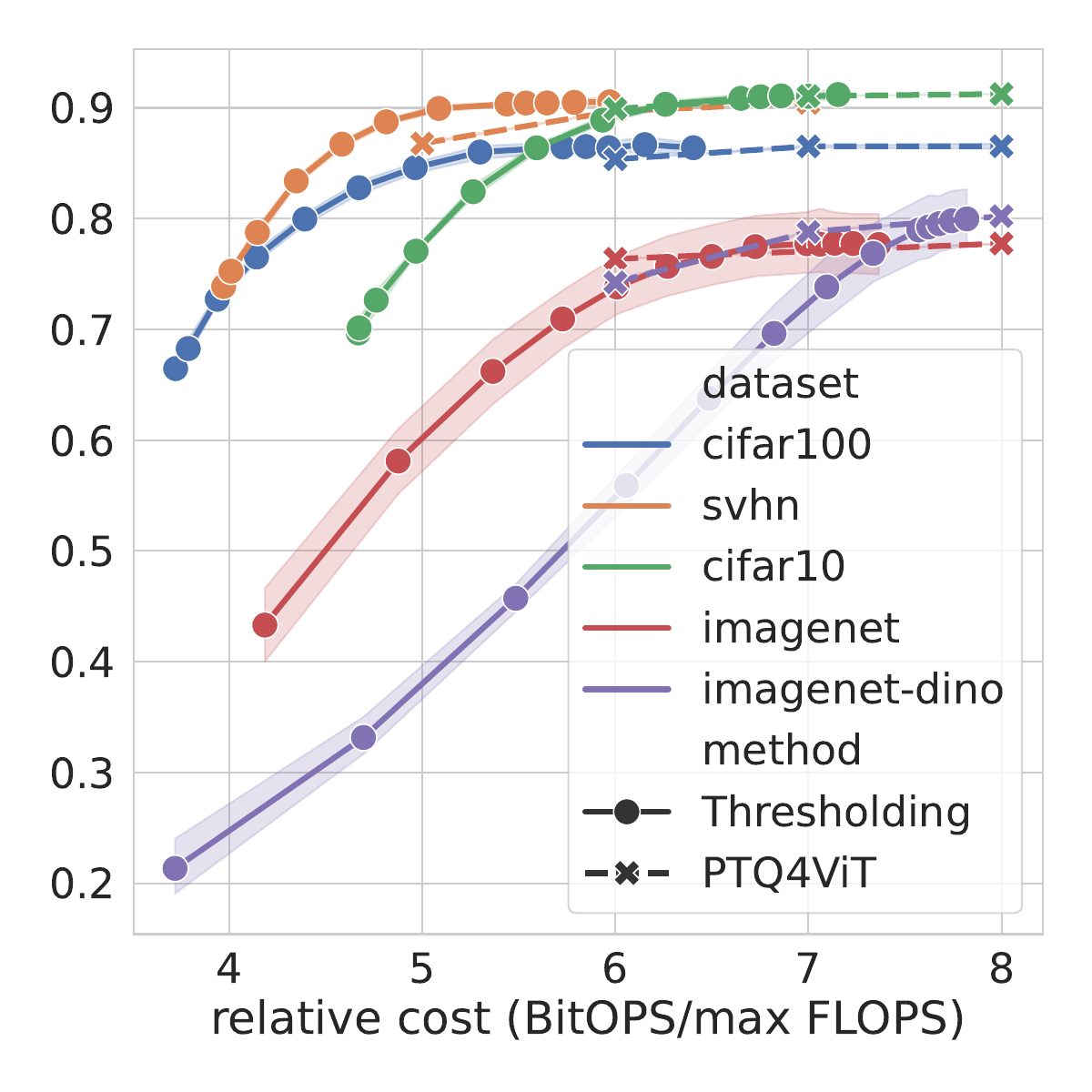}};
  \node[black] at (-1.5,1.7) {C)};
\end{tikzpicture}
\end{minipage}
    \begin{minipage}{0.325\linewidth}
     \begin{tikzpicture}
  \node[inner sep=0pt] (A) { \includegraphics[scale=0.23]{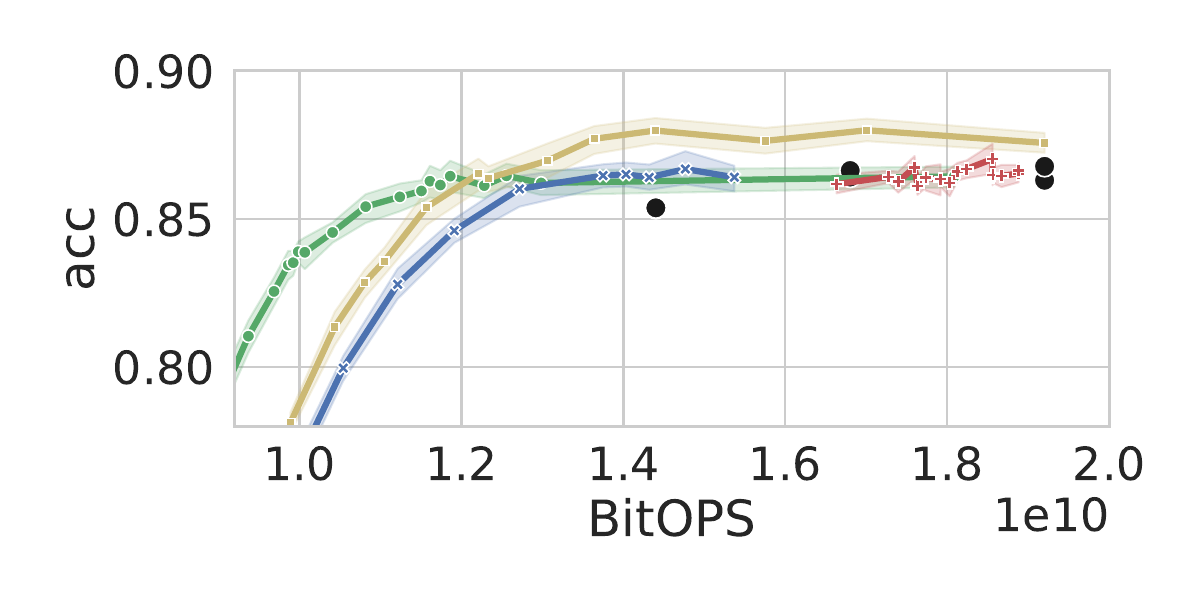}};
  \node[black] at (1.7,-0.3) {D)};
\end{tikzpicture}
    \end{minipage}
    \begin{minipage}{0.325\linewidth}
         \centering
          \begin{tikzpicture}
  \node[inner sep=0pt] (A) { \includegraphics[scale=0.23]{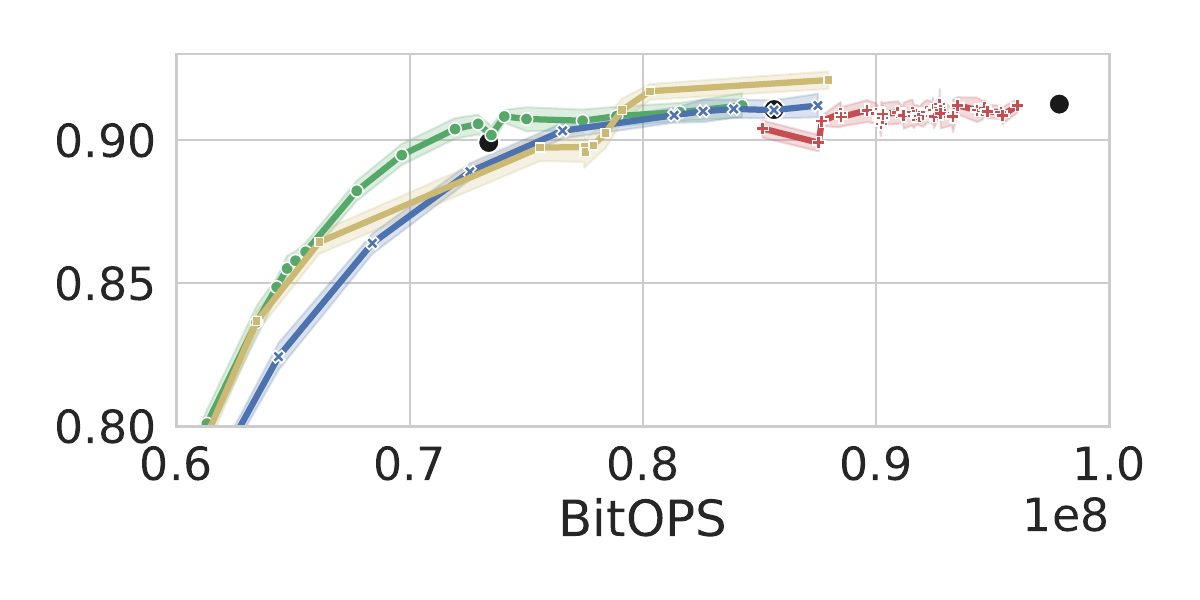}};
  \node[black] at (1.7,-0.3) {E)};
\end{tikzpicture}
    \end{minipage}
    \begin{minipage}{0.325\linewidth}
         \centering
    \begin{tikzpicture}
  \node[inner sep=0pt] (A) { \includegraphics[scale=0.23]{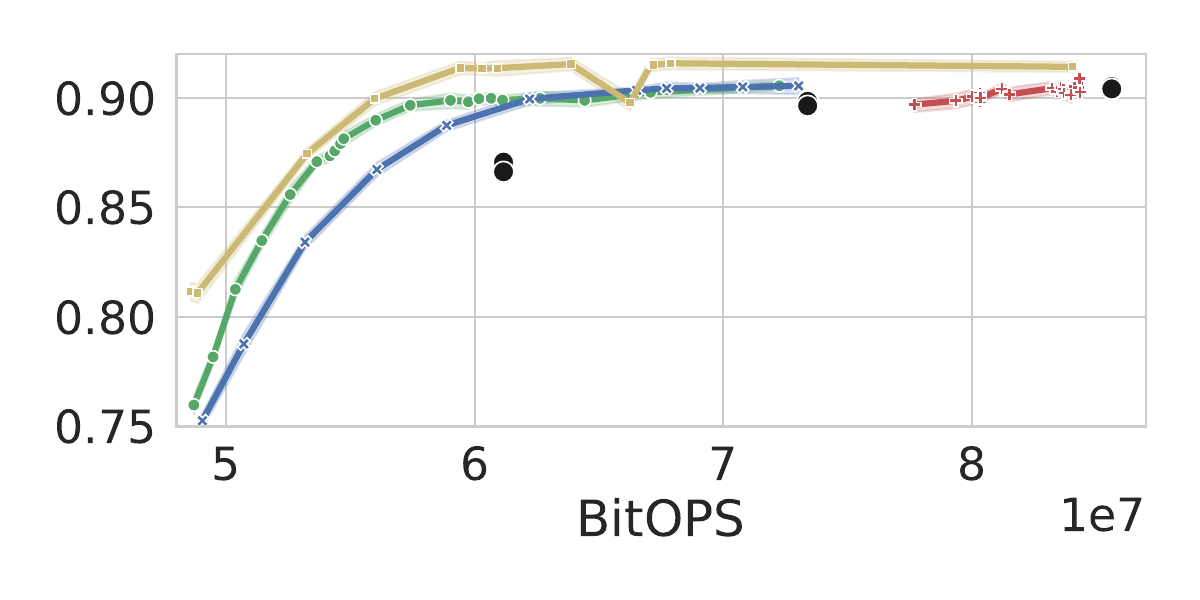}};
  \node[black] at (1.9,-0.3) {F)};
\end{tikzpicture}
    \end{minipage}
      \caption{Accuracy vs BitFLOPS.\textbf{ A)} Imagenet t2t-Vit-14, \textbf{B)} Imagenet Vit-14 with DinoV2,  \textbf{C)} EE vs quantization performance, \textbf{D)} CIFAR100 t2t-Vit-14, \textbf{E)} CIFAR10 t2t-Vit-7, \textbf{F)} SVHN t2t-Vit-7. }
\label{fig:result_bas}\vspace{-1em}
\end{figure}

The performance ordering of each baseline is maintained across datasets and backbone architectures, as presented in Figure~\ref{fig:result_bas}. Overall, QuEE outperforms the baselines, especially for lower-cost regimes, with the exception of the SVHN dataset, which we will discuss later. JEIDNN closely follows, performing particularly well in higher-cost regimes, where it outperforms both QuEE and the initial accuracy of the backbone due to its ability to train the IMs jointly with the gating mechanism.  DQNet is less competitive, as it is limited in the range of costs it can reach, since it cannot early exit. It generally outperforms simple quantization (PTQ4ViT), maintaining the same accuracy at a lower cost. We emphasize that both JEIDNN and DQNet require retraining, with adjustment of the trade-off hyperparameters, to obtain each point along the operating curve. This leads to instabilities that are apparent in the results, making these methods less reliable. In practice, this means that in order to obtain a specific target cost or performance, the models would need to be retrained many times until the desired values are attained. In contrast, QuEE only has to be trained once to obtain a full curve, because the cost can be modified at inference to obtain a different trade-off. This makes QuEE more efficient, stable, and flexible, and therefore more practical.

As we stated in our methodology, the strength of QuEE lies in its ability to combine quantization with early exit. For this to be an advantageous strategy, both computation reduction methods need to perform relatively well independently — there is no benefit in combining a weak method with a strong one. As a result, we expect QuEE to perform the best when both quantization and early exit are approximately equally effective. This is confirmed by our experimental results. Figure~\ref{fig:result_bas}C) compares both computation reduction methods for the various experiments. When the methods are on par with each other, as is the case for  CIFAR-10, ImageNet with T2T-ViT, and CIFAR-100, QuEE obtains its best performance. In contrast, for SVHN, the quantization methods are significantly outperformed by early exit, and the early exit method JEIDNN outperforms QuEE on this dataset. SVHN is a relatively simple dataset, and the T2T-ViT-7 architecture is much deeper than necessary for this task, making it particularly suitable for early exit. Conversely, for ImageNet with ViT pre-trained with Dino, we observe the inverse trend, and quantization methods outperform early exit. This indicates that the intermediate representations of ViT are not well suited to be used as input to perform inference, which could be a side effect of the self-supervised objective in  DinoV2~\cite{oquab2024dinov2}. This could also explain the instabilities of JEIDNN, because it jointly trains the inference heads with the gating mechanisms.

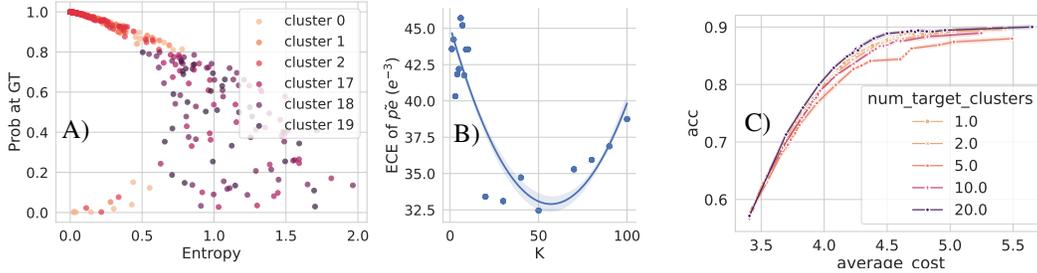
\begin{figure}
\begin{minipage}[t]{0.32\linewidth}
         \centering
          \begin{tikzpicture}
  \node[inner sep=0pt] (A) {  \includegraphics[scale=0.3]{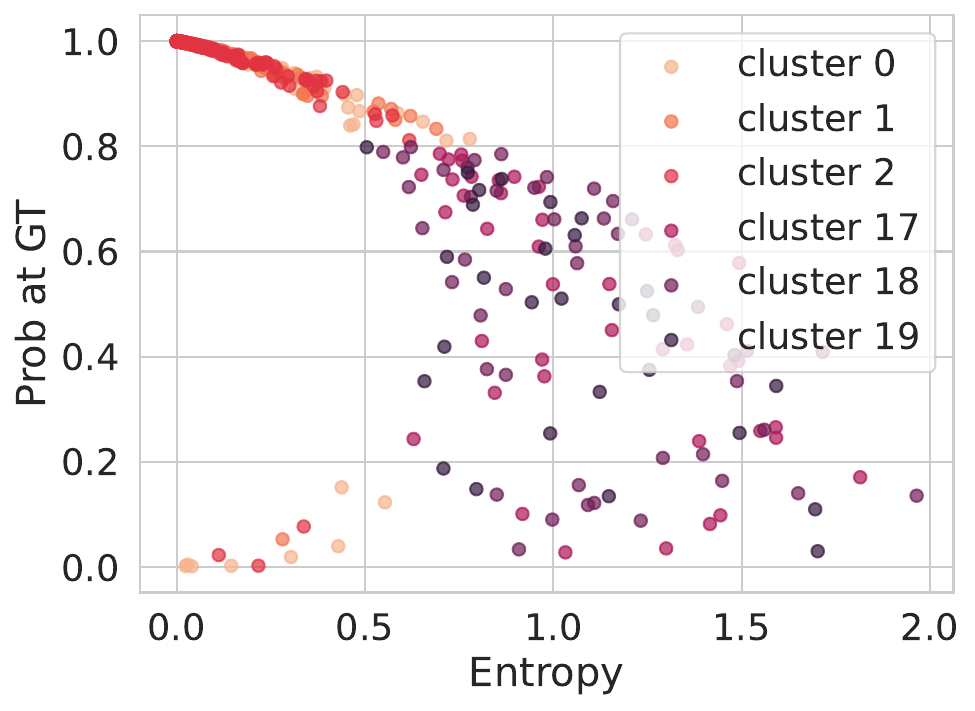}};
  \node[black] at (-1.5,0) {A)};
\end{tikzpicture}
    \end{minipage}
    \begin{minipage}[t]{0.32\linewidth}
   \centering
    \begin{tikzpicture}
  \node[inner sep=0pt] (A) { \includegraphics[scale=0.3]{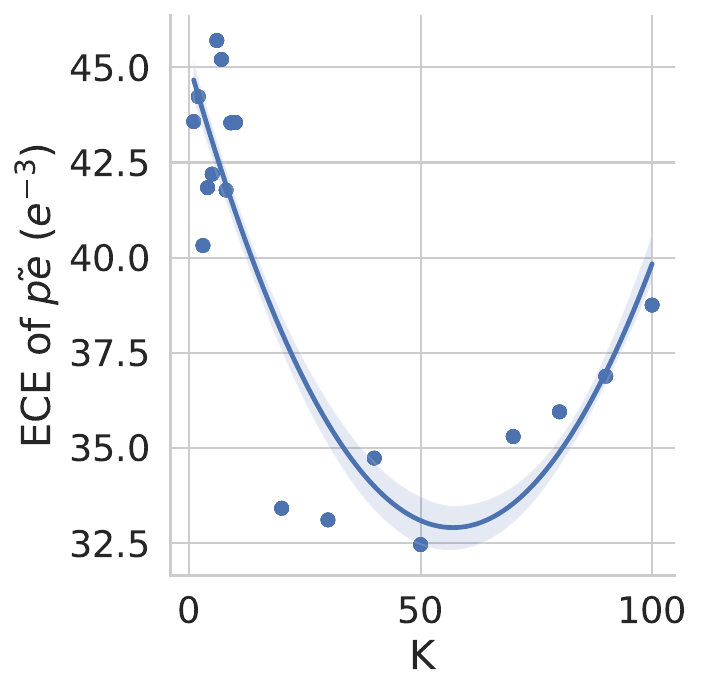}};
  \node[black] at (-0.6,0) {B)};
\end{tikzpicture}

    \end{minipage}
      \begin{tikzpicture}
  \node[inner sep=0pt] (A) { \includegraphics[scale=0.3]{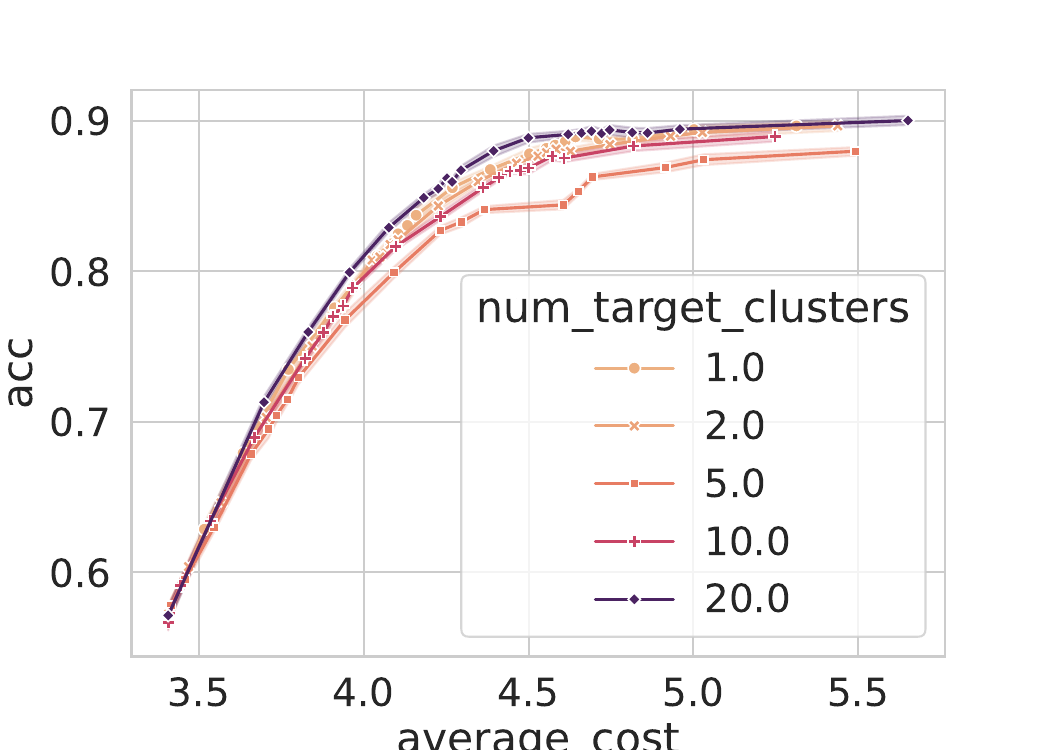}};
  \node[black] at (-1.7,0) {C)};
\end{tikzpicture}

    \caption{\textbf{A)} Predicted probability of the ground truth label and entropy of a few clusters for $K=20$ on the SVHN datasets. We can see that there is a correlation between the clustering and those two metrics. \textbf{B)} Expected calibration error of $\tilde{pe}$ generated from varying cluster numbers $K$ for the SVHN dataset on T2T-ViT-7 with a fitted second order polynomial. Increasing the number of clusters (up to 50) reduces the calibration error. We discuss this further in Appendix \ref{app:cluster} \textbf{C)} Accuracy-cost trade-off for different values of $K$ for the SVHN dataset on T2T-ViT-7.}\label{fig:cluster}\vspace{-1em}
\end{figure}

\newpage
\paragraph{Accurately predicting probability of errors leads to better performance for high cost regime}
\begin{wrapfigure}{r}{0.4\textwidth}\vspace{0em}
    \centering
    \includegraphics[scale=0.3]{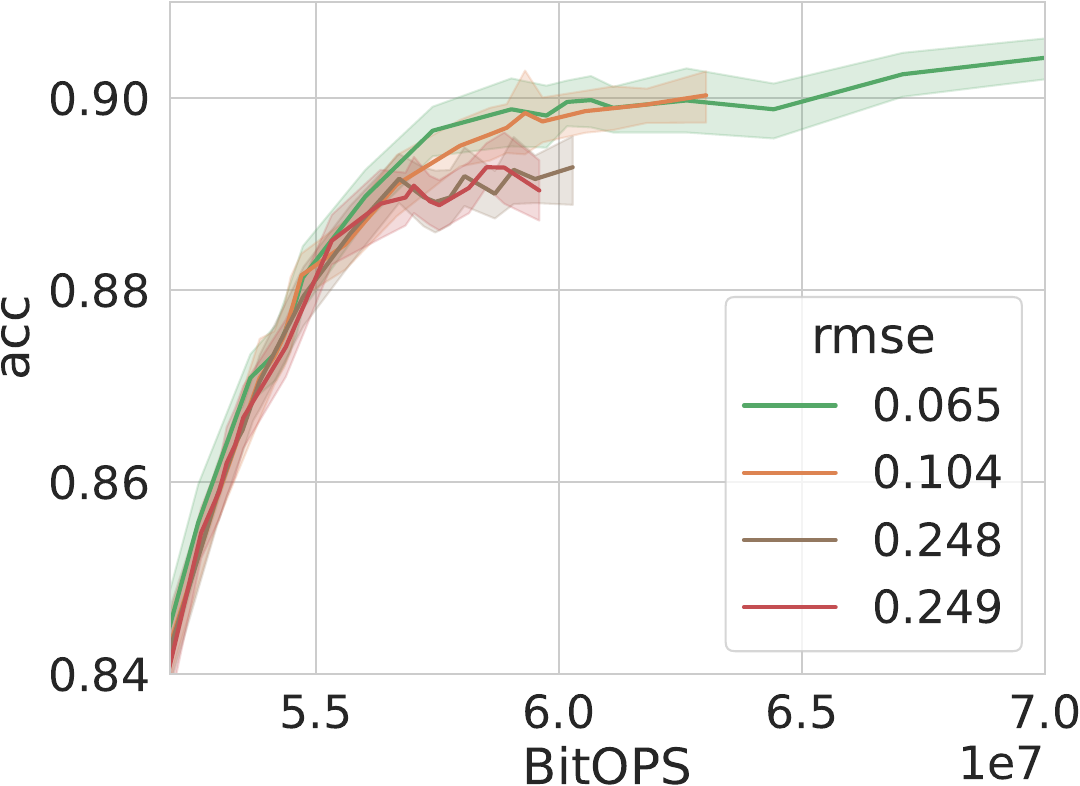}
    \caption{Accuracy vs cost curves of predictors with varying RMSE performance on the SVHN dataset. } \vspace{-1em}
    \label{fig:rmse}
\end{wrapfigure}
Our algorithm involves prediction of the approximated probabilities of errors $ \widetilde{pe}(f_{\pi}  |x_i)  $, as defined in Eqn~\ref{eq:approx}. We assess the accuracy of these predictions by calculating the root mean square error on test samples $\textrm{rmse} = \| \widetilde{pe}(f_{\pi}  |x_i) -h^j_{\theta} \|^2_2$, and analyze how prediction accuracy correlates with performance on the primary task.  In Figure~\ref{fig:rmse}, we present the accuracy/cost curves of our QuEE algorithm for different hyperparameters on one dataset. As expected, lower RMSE results in better performance. Moreover, the gap in performance is notable in the higher cost regime. This outcome is a direct consequence of the fact that the gating decisions are based on $ c_{\pi} + \hat{pe}_{\theta}(f_{\pi}|x)$. In a low-cost regime, the $c_{\pi}$ are given more relative weight, and the $\hat{pe}_{\theta}(f_{\pi}|x)$ have less influence on the gating decision. Consequently, the overall performance is less affected by poor predictions. However, in a high-cost regime, the $\hat{pe}_{\theta}(f_{\pi}|x)$ become more important and we can observe a widening performance gap. QuEE with poor predictors cannot even reach higher cost operating points as it struggles to learn to use the more costly classifiers.
\paragraph{Clustering analysis}
In this section, we verify that the choices we made to construct our approximation of the probability of error $\tilde{pe}$ are sensible, and that they lead to a reasonable approximation. First, to validate our approach, we observe the grouping capability of the clusters for certain metrics that are likely related to the ground truth probability of error: the predicted probability of the ground truth class $\pred_y$ and the entropy of $\pred$. We collect these metrics for test samples with the cluster assigned to each for various $K$. In Figure~\ref{fig:cluster}, we can see that some clusters (clusters 0, 1, 2) contain samples that predominantly have low entropy and high predicted $\pred_y$, while other clusters (clusters 17, 18, 19) consist almost entirely of samples with relatively high entropy (>0.5) and lower $\pred_y$. Next, in Figure~\ref{fig:cluster} \textbf{B)}, we compute the calibration error for our assigned $\tilde{pe}$ on test samples. On average, increasing the number of clusters $K$ improves the calibration error until we reach $K=50$. Beyond this, a deteriorating calibration performance is presumably caused by the sampling error. Lastly, in Figure~\ref{fig:cluster} \textbf{C)}, we can also observe how $K$ affects the efficiency performance. For $K=1$, the prediction task is very easy -— the modules only need to learn to predict a single fixed value for each path. Therefore, we see a decrease of performance as we increase $K$ at first, reaching its lowest at $K=5$; then the accuracy of the target starts to improve, and we ultimately reach the best performance using larger $K$.

\section{Conclusions and Limitations}

We have introduced a post-training method that can adaptively combine different computation reduction methods such as quantization and early exiting. For future work, we will consider integrating other post training computation reduction methods that were not included in this work, such as pruning or distillation. We will also extend the approach to allow for joint training of the inference heads alongside our gating mechanism, as in~\cite{regol2024_jeidnn}.

\textbf{Limitations.} The main limitation of our approach is scaling; the algorithm has an inference complexity of $O(B^EE)$. Although we have found that this is actually not that prohibitive in practice as the relative cost of each gate is extremely small compared to the actual network, such complexity can obviously be a scaling issue. We found that a simple sampling approach can mitigate the issue without significantly impacting the accuracy-cost trade-off. A second limitation of our work is the cost computation. The BitOPS metric that we are using can be viewed as a theoretical metric. The real efficiency benefit in terms of speed and memory remains untested. The practical application of quantized networks requires careful handling and a dedicated kernel implementation. We view this as beyond the scope of this work, but it is clearly an important aspect to tackle for implementation of our proposal in practice. Finally, constructing an approximation of the target $PE$ for training the decision modules relies on unverifiable assumptions. We offer experimental analysis indicating the reasonableness of our approximation. However, this is not a significant limitation, as it serves merely as an intermediate step toward our main objective, which is measurable: efficiency improvement.

\bibliographystyle{IEEEtranSN}
\bibliography{ref}

\clearpage
\section{Appendix}
\begin{table}[h]
    \centering\caption{Notation table} \vspace{1em}
    \begin{tabular}{cl} \toprule 
    \textbf{symbol} & \textbf{description} \\ \midrule
     $\mathcal{X}$  & Input space.\\
     $\mathcal{Y} = \{1, 2, \dots, |\mathcal{Y}|\}$  & Label space.\\
      $P(Y|X)$  & ground truth conditional distribution. \\
      $PE(f(x)|x)  $  & Probability of making an error with prediction $y$ given $x$. equal to $1-\Pr(Y=y|X)$ \\ \midrule
     \multicolumn{2}{c}{\textbf{General dynamic network notation}}  \\  \midrule  
     $f_l$  &  The $l$-th classifier.\\
      $c_l$  &  The cost associated to the $l$-th classifier.\\
     $S(x) \in [L]$  &  A classifier-selector function (with negligible evaluation cost compared to $c_l$).\\
       \midrule  
      \multicolumn{2}{c}{\textbf{QuEE notation}} \\
      \midrule 
      $E$  &  Number of early exits in the network.\\
        $B$ & Number of levels of computation (quantization) that can be used per layer.\\
         $L = \sum^E_{e=1}B^e$ & Total number of classifiers/paths that we can select.\\
        $\pi = b_1 \to \dots \to  b_e$ & A path.  A path contains the level of computation used by each layer. \\
        $\mathcal{P}$ & The set of all paths. $|\mathcal{P}| = L$.\\
        $\mathcal{P}_{\pi}$ & The set of all paths starting with $\pi$.\\
        $f_{\pi} : \mathcal{X} \to \mathcal{Y}$ & The classifier corresponding to $\pi$.   \\
         $c_{\pi} \in [0,1]$ & The cost of evaluating $f_{\pi}$.   \\
         $\pred_{\pi} \in [0,1]^{|\mathcal{Y}|} $ & The predicted prob. vector of $f_{\pi}$.   \\
          \midrule 
         $g^e_{\theta}(\cdot) $ & Gate placed at exit $e$. Decides on whether we should exit or use computation level $b$.   \\ \bottomrule
    \end{tabular}
    
    \label{tab:notation}
\end{table}
\subsection{Experiment details}\label{app:exp_details}
\subsubsection{Backbones}
In our experiments we used the T2T-ViT \cite{yuan2021_t2t_vit} and ViT \cite{vit} vision transformers pre-trained with DinoV2 \cite{oquab2024dinov2}. We provide now more details about these two architectures.

\paragraph{ViT with DinoV2} The vision transformer proposed by \citeauthor{vit} splits an input image into patches which are projected as embeddings and combined with positional encoding \cite{vit}. The architecture also uses a special ``classification'' token that is fed as input to a classification head at the last layer. The embedding size is set as a hyper-paramter. We use ViT-S-14 throughout our experiments. S (``small'') corresponds to an embedding size of 384 while 14 is the patch size. We use a 12-layer ViT in all our experiments. The ViT backbones were all pre-trained using the DinoV2 procedure \cite{oquab2024dinov2} which uses self-supervision on a variety of large datasets to train a foundation model that can be used on downstream vision tasks with minimal fine-tuning \cite{oquab2024dinov2}. Specifically, the backbone of the foundation model does not need to be retrained and only the inference head needs to be adapted to the dataset and task at hand \cite{oquab2024dinov2}.
\paragraph{T2T-ViT} The token-to-token vision transformer is a more data-efficient vision transformer architecture \cite{yuan2021_t2t_vit}. T2T-ViT uses a progressive tokenization scheme to fuse neighbouring embeddings into a single embedding at the input layer, capturing local structure and reducing the token size. This greatly reduces the number of trainable parameter, making the model more data-efficient \cite{yuan2021_t2t_vit}. In our experiments we used the 7-layer and 14-layer T2T-ViT models. \citeauthor{yuan2021_t2t_vit} provide the parameters for these 2 architectures pre-trained on ImageNet \cite{imagenet} which we used as-is for ImageNet experiments on T2T-ViT. For other datasets, we use the transfer-learning procedure also provided by \citeauthor{yuan2021_t2t_vit} to fine-tune the backbone.

\subsubsection{Baselines}\label{app:baselines}
In this section we discuss the different baselines we use in greater depth.

\paragraph{PTQ4ViT} PTQ4ViT is a SOTA post-training quantization algorithm specialized for vision transformers \cite{PTQ4ViT}. \citeauthor{PTQ4ViT} observe that the activations in vision transformers do not exhibit the usual Gaussian distributions reported in CNN-based models. For example, the values after the GeLU cover a very large positive range and a restricted negative range. To address this, they propose a twin-uniform quantization scheme where uniform quantization is applied after splitting the activations in two disjoint regions (positive and negative). This allows to find optimal quantization parameters for each region separately \cite{PTQ4ViT}. The proposed algorithm uses the Hessian of the loss with respect to the parameters as an indicator of sensitivity to determine the optimal quantization parameters \cite{PTQ4ViT}. We obtain the different operating points by quantizating the entire network with PTQ4ViT at different bit resolutions.

\paragraph{DQNet} DQNet proposes a data-adaptive mixed-precision quantization scheme for image classification \cite{Liu2022_dqnet}. A lightweight bit-controller module is attached to the network to be quantized. The bit-controller's role is to determine the optimal bit width of each layer for a given sample \cite{Liu2022_dqnet}. It is placed at a user-specified layer and uses the feature map output by that layer to determine the optimal bit path. The DQNet data-adaptive quantization procedure can be applied to any architecture and is compatible with other  quantization schemes \cite{Liu2022_dqnet} making it a perfect candidate for our setup. For fairer comparison, we integrate DQNet with PTQ4ViT\cite{PTQ4ViT} which provides better quantization results on vision transformers. As such, we slightly modify the training procedure to work in a post-training way, in line with our pre-trained model setting. The training procedure used first quantizes the trained network using PTQ4ViT and then trains the bit-controller separately. The algorithm uses a cost-aware loss which allows the user to emphasize efficiency or accuracy. This is done by introducing a bit-loss term $\mathcal{L}_{\text{bit}} = \sum_{l = 1}^L b_l - b_{\text{target}}$ which measures the distance of the selected bit widths to a user-specified target bit \cite{Liu2022_dqnet}. The overall DQNet loss is $\mathcal{L} = \mathcal{L}_{\text{CE}} + \alpha \mathcal{L}_{\text{bit}}$ where $\alpha$ is a hyper-parameter. We obtain various operating points by varying the target bit width as well as $\alpha$.

\paragraph{JEIDNN} JEIDNN is an early exit framework that is compatible with frozen backbones. \citeauthor{regol2024_jeidnn} augment a pre-trained backbone network with classifiers and controllers. The role of the controllers is to determine whether a sample can be exited or should be propagated along the network for further processing. The controllers and classifiers are jointly trained to allow for the ``specialization'' of the classifiers on points that they will effectively handle at inference time \citet{regol2024_jeidnn}. JEIDNN also uses a cost-aware loss, controlled by a hyper-parameter $\lambda$ which encodes the cost-accuracy trade-off. We obtain different operating points in our experiments by running JEIDNN with varying $\lambda$ values.

\subsubsection{Datasets}\label{sec:app_dataset}
\paragraph{ImageNet} consists of 1.2 million training images representing 1,000 balanced classes \cite{imagenet}. The images are resized to $256\times 256$ pixels. We use 1.1 million images for training as well as 50,000 images as a validation set and 50,000 images for testing. The validation set is used for early-stopping as well as hyper-parameter grid-search. We also use 600 images for generating the K-means clusters use to discretize the predictors $h_{\theta}^j$ training targets.
\paragraph{CIFAR} \cite{krizhevsky_cifar} datasets consist of 60,000 small $32\times32$ coloured images. CIFAR10 consists of 10 classes while CIFAR100 spans 100 classes. CIFAR10 images were resized to $70\times70$ pixels when used with ViT-DinoV2 \cite{vit, oquab2024dinov2} models (the image size needs to be a multiple of the patch size of 14) and $64\times64$ for T2T-ViT  models \cite{yuan2021_t2t_vit}. CIFAR100 images were resized to $224\times224$ for T2T-ViT following \citet{yuan2021_t2t_vit} and $70\times70$ for Vit-DinoV2. We reserve 5,000 images for the validation set (hyper-parameter search, early-stopping and discretization of targets) and 10,000 images for testing.

\paragraph{SVHN} SVHN~\citep{svhn} dataset consists of 73,257 training images and 26,032 testing images. We reserve 5,000 images from the training set as validation data. The images are $32 \times 32$ coloured images of house numbers. The version of SVHN we use spans 10 classes where the label corresponds to the central digit of the house number while the remaining digits are noise.

The results reported correspond to 95$\%$ confidence intervals on the means of our results. These intervals are obtained by using a boostrap procedure of 10 trials where the test set is split into 10 subsets.

\subsubsection{Experiment parameters}\label{app:exp_parameters}
We share experiment parameters in this section, describing each step of the experiment in a paragraph. Table \ref{tab:common_hyperparm} summarizes shared hyper-parameters for all algorithms (QuEE and baselines). The highest bit resolution was chosen as the smallest bit width that maintains the overall network accuracy. The lower resolutions were chosen so that the performance drop of the last prediction head never exceeds 10\%.

\paragraph{Intermediate classifiers training}
For algorithms that use early-exiting, we augment the pre-trained frozen network with intermediate inference modules (IMs) are user-specified layers. This step is skipped for quantization-only algorithms (DQNet \cite{Liu2022_dqnet}, PTQ4ViT\cite{PTQ4ViT}) Table \ref{tab:common_hyperparm} shows the exits used for each dataset-architecture pair. These IMs are trained jointly until convergence using a simple scaled loss as was done in \citet{regol2024_jeidnn}. Table \ref{tab:common_hyperparm} indicates the maximum number of IM training which we use in conjunction with early-stopping. In practice SVHN converges at around 7 epochs, CIFAR10 at 10, CIFAR100 at 13 and ImageNet at around 15. We use the ADAM optimizer \cite{adam_opt} with an initial learning rate of $1e^{-3}$ and a weight-decay of $1e^{-4}$. We note that the backbone parameters are frozen as these backbones have been trained extensively on large datasets \cite{oquab2024dinov2, loftq_lora}. This is well-aligned with the setup of working with foundation models where only the inference head needs to be optimized on a specific dataset. 

\paragraph{Quantization} We quantize the pre-trained backbones using PTQ4ViT \cite{PTQ4ViT}. As a post-training quantization algorithm, it requires a small amount of data for calibration. We use 128 calibration samples for all datasets and all models. This is sufficient as \citet{PTQ4ViT} achieve SOTA quantization using only 32 images. The bit resolution per dataset are summarized in Table \ref{tab:common_hyperparm}. For early-exiting baselines (JEIDNN and EE with thresholding), we use the largest bit-width for each dataset (8 for all datasets but SVHN where we used 7). This is to ensure a fair comparison.

\paragraph{Algorithm-specific information} 
\begin{itemize}
    \item \textbf{QuEE} Starting from a multi-quantization network with trained IMs we train the QuEE predictors on the discretized probability of error (obtained via K-means clustering on the predicted probability $\pred_{\pi}$ for each path $\pi$). Table \ref{tab:quee_hyperparams} summarized QuEE-specific hyper-parameters. The number of clusters was chosen based on the ECE using $\widetilde{pe}$ computed on the validation set computed on 600 samples from the validation set. We used \href{https://scikit-learn.org/stable/modules/generated/sklearn.cluster.KMeans.html}{SciKit learn's K-Means} algorithm using the L2 norm as a distance metric where the clusters centroids are randomly initialized 10 times and the run with the lowest inertia is kept. We then train the gates using the ADAM optimizer \cite{adam_opt} with an initial learning rate of $1e^{-3}$ and a weight-decay of $1e^{-4}$.
    \item \textbf{DQNet} As mentioned earlier, we use PTQ4ViT\cite{PTQ4ViT} as a quantization algorithm on the DQNet \cite{Liu2022_dqnet} architecture since it is compatible with any quantization algorithm. We place the bit-controller at the earliest exit as indicated in Table \ref{tab:common_hyperparm}. We iterate over the bit resolutions listed in Table \ref{tab:common_hyperparm} while sweeping values of $\alpha \in \{0, 0.01, 0.05, 0.1, 0.3, 0.5, 2, 10\}$. We use the same number of training epochs as QuEE (see Table \ref{tab:quee_hyperparams}).  We also use the ADAM optimizer \cite{adam_opt} with an initial learning rate of $1e^{-3}$ and a weight-decay of $1e^{-4}$.
    \item \textbf{PTQ4ViT} We obtain different operating points for the accuracy-cost curves using PTQ4Vit \cite{PTQ4ViT} by setting the bit resolution to each bit resolution listed in Table \ref{tab:common_hyperparm}.
    \item \textbf{JEIDNN} As a base JEIDNN network we use the highest bit resolution listed in Table \ref{tab:common_hyperparm} for each dataset. We augment the network with gating modules following the structure of the original algorithm \cite{regol2024_jeidnn}. We use a bi-level batch count of 200 unless the training data loader had less than 200 batches in which case we set the bi-level batch count to half the number of batches in the data loader. We obtain different operating points by sweeping the cost-accuracy trade-off hyper-parameter $\lambda \in \{0, 0.01, 0.05, 0.1, 0.3, 0.5, 0.8, 1, 1.5, 2, 2.5, 3, 5\}$. We also use the ADAM optimizer \cite{adam_opt} with an initial learning rate of $1e^{-3}$ and a weight-decay of $1e^{-4}$.
    
\end{itemize}

\begin{table}[h!]
\centering\caption{Hyper-parameters for IM pre-training and quantization for all algorithms} \vspace{1em}
\begin{tabular}{ccccccc}
\toprule
\textbf{Dataset} & \textbf{Arch} & \textbf{\begin{tabular}[c]{@{}c@{}}Input \\ size\end{tabular}} & \textbf{\begin{tabular}[c]{@{}c@{}}IM\\ train.\\ max. epoch\end{tabular}} & \textbf{\begin{tabular}[c]{@{}c@{}}Bit\\ res.\end{tabular}} & \textbf{\begin{tabular}[c]{@{}c@{}}Batch \\ size\end{tabular}} & \textbf{Exits}\\
\midrule
SVHN & T2T-ViT-7 & 32x32 & 10 & 5,6,7 & 512 & 2,4,7 \\
CIFAR10 & T2T-ViT-7 & 64x64 & 15 & 5,6,7,8 & 512 & 2,4,7\\
CIFAR100 & T2T-ViT-14 & 224x224 & 20 & 6,8 & 128 & 5,7,11,14\\
ImageNet & T2T-ViT-14 & 224x224 & 20 & 6,8 & 128 & 5,7,11,14\\
ImageNet & ViTs14 & 256x256 & 20 & 6,7,8 & 128 & \begin{tabular}[c]{@{}c@{}}4,6,8,\\ 10,12\end{tabular}\\
\bottomrule
\end{tabular}\label{tab:common_hyperparm}
\end{table}

\begin{table}[h!]
\centering\caption{Hyper-parameters for QuEE} \vspace{1em}
\begin{tabular}{ccccccccc}
\toprule
\textbf{Dataset} & \textbf{Arch} & \textbf{\begin{tabular}[c]{@{}c@{}}Num\\ clusters \\ (K)\end{tabular}} & \textbf{\begin{tabular}[c]{@{}c@{}}Batch\\ size\end{tabular}} & \textbf{\begin{tabular}[c]{@{}c@{}}QuEE\\ training\\ epochs\end{tabular}} & \textbf{\begin{tabular}[c]{@{}c@{}}Emb.\\ size\end{tabular}} & \textbf{\begin{tabular}[c]{@{}c@{}}Hid.\\ dim. \\ size\end{tabular}}  & \textbf{\begin{tabular}[c]{@{}c@{}}Avg. training time\\ NVIDIA GeForce  \\ GTX 2070\end{tabular}} \\
\midrule
SVHN & T2T-ViT-7 & 70 & 512 & 10 & 8& 16 & 5 min.\\
CIFAR10 & T2T-ViT-7 & 20 & 512 & 10 & 8 & 16 & 13 min.\\
CIFAR100 & T2T-ViT-14 & 70 & 128 & 20 &8 & 32 & 1.6 hour\\
ImageNet & ViTs14 & 40 & 128 & 20 & 8 & 32 & 4.5 hour \\
\bottomrule
\end{tabular}\label{tab:quee_hyperparams}
\end{table}

\subsection{Additional clustering discussion}\label{app:cluster}
As mentioned in Section \ref{sec:approximating_prob_error}, we assume that the probability of error $PE(f_{\pi}|x_i)$ is smooth over a transformation of the predicted probability space $\pred_{\pi}(x)$ for a given path $\pi$. We thus discretize the space into K partitions denoted $\{Q_1, \dots, Q_K\}$ and compute a ``delegate'' value $\widetilde{pe}(f_{\pi, Q})$ for each cluster Q. During training of the predictor modules when 

We now provide additional experimental results for the clustering of target probabilities of error discussed in in Section \ref{sec:approximating_prob_error}.

\begin{figure}[h!]
    \centering
    \begin{minipage}{0.48\linewidth}
    \includegraphics[scale=0.45]{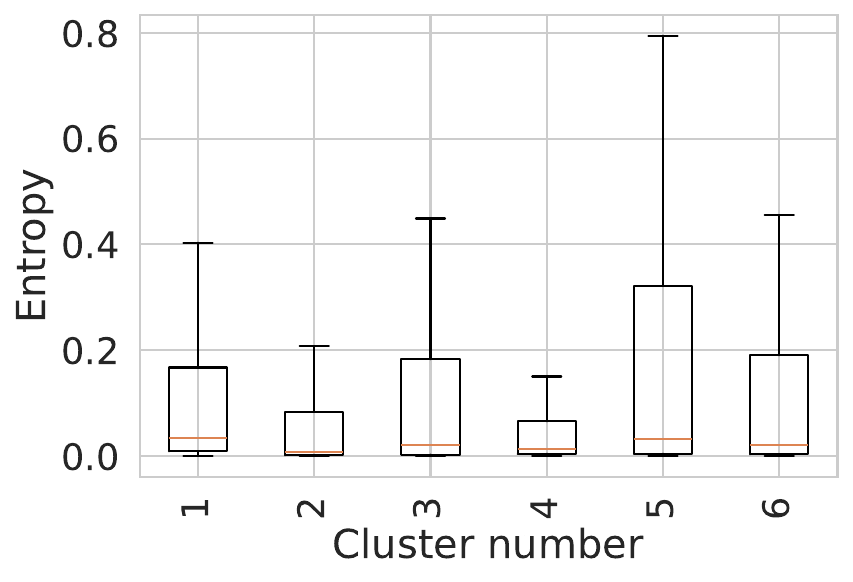}
    \end{minipage}
    \begin{minipage}{0.48\linewidth}
         \centering
    \includegraphics[scale=0.45]{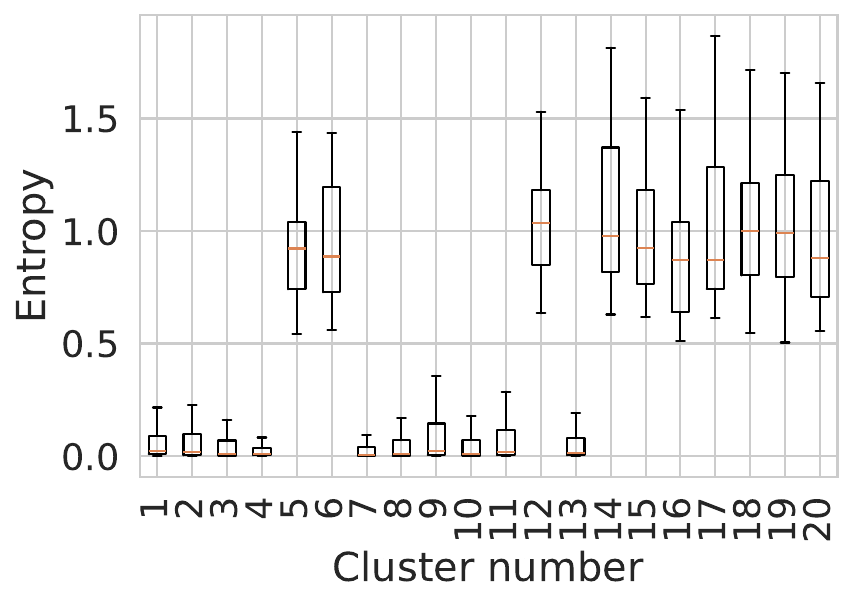}
    \end{minipage}
      \caption{Box-plots showing the entropy of each cluster computed for SVHN on T2T-ViT-7 at path $\pi = 7 \rightarrow 6 \rightarrow 6$ with K=6 (left) and K=20 (right).}\label{fig:cluster_analysis_svhn_entropy_box_plot}
\end{figure}

\begin{figure}[h!]
    \centering
    \begin{minipage}{0.48\linewidth}
    \includegraphics[scale=0.45]{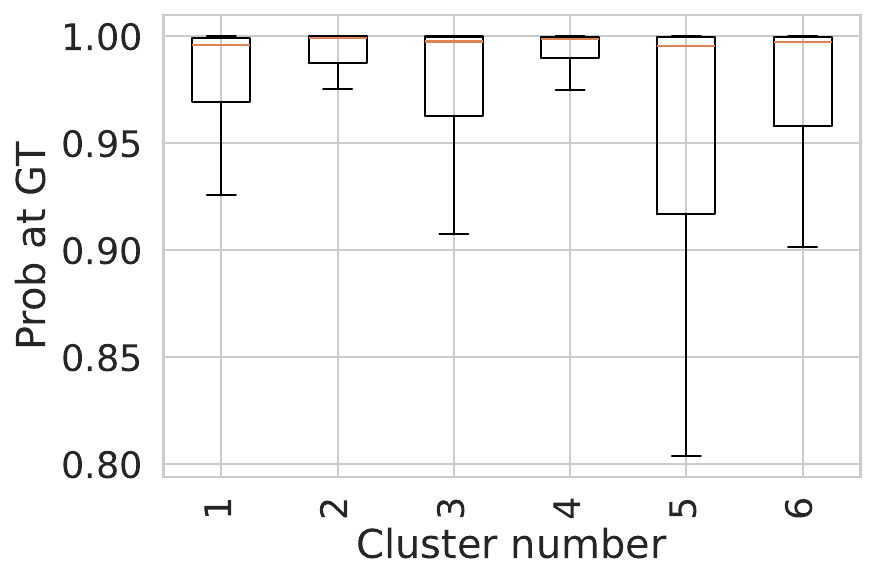}
    \end{minipage}
    \begin{minipage}{0.48\linewidth}
         \centering
    \includegraphics[scale=0.3]{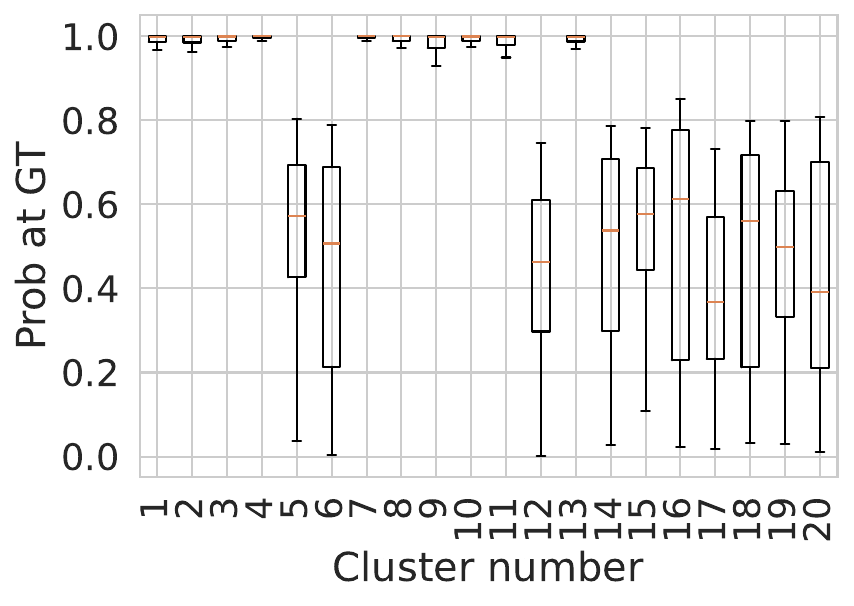}
    \end{minipage}
      \caption{Box-plots showing the predicted probabilty at the ground-truth class of each cluster computed for SVHN on T2T-ViT-7 at path $\pi = 7 \rightarrow 6 \rightarrow 6$ with K=6 (left) and K=20 (right).}\label{fig:cluster_analysis_svhn_prob_gt_box_plot}
\end{figure}

\begin{figure}
    \centering
    \includegraphics[scale=0.6]{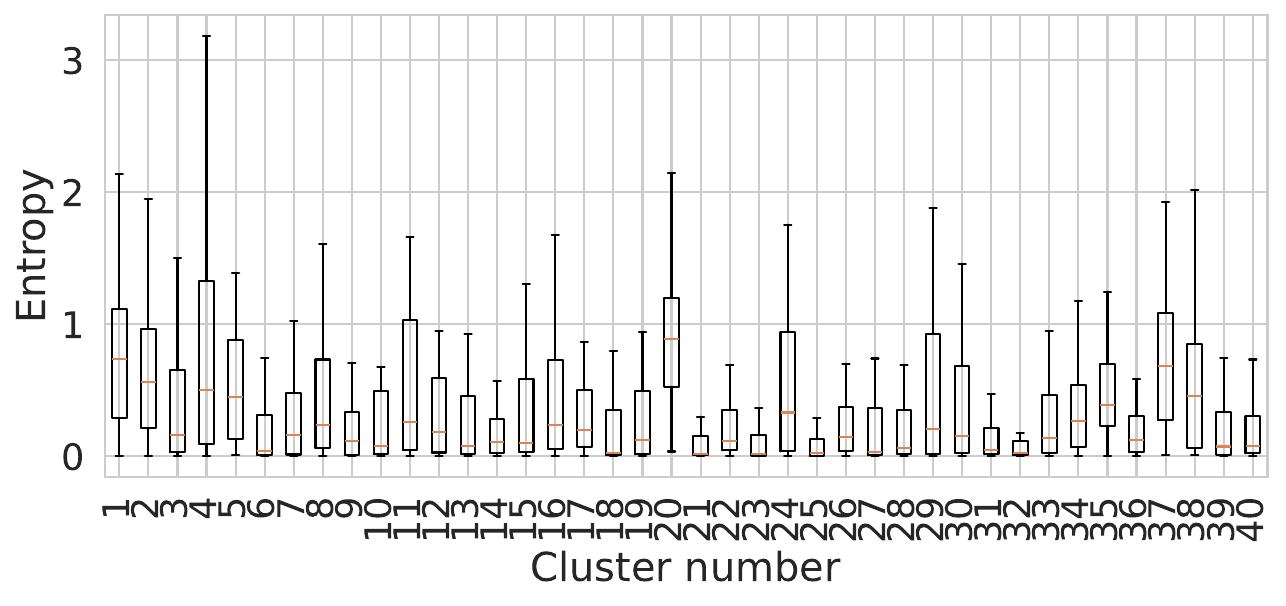}
    \caption{Box-plots showing the entropy of each cluster computed for CIFAR100 on T2T-ViT-14 at path $\pi = 8 \rightarrow 7$ with K=40}
    \label{fig:cifar100_k_40}
\end{figure}

\begin{figure}
    \centering
    \includegraphics[scale=0.6]{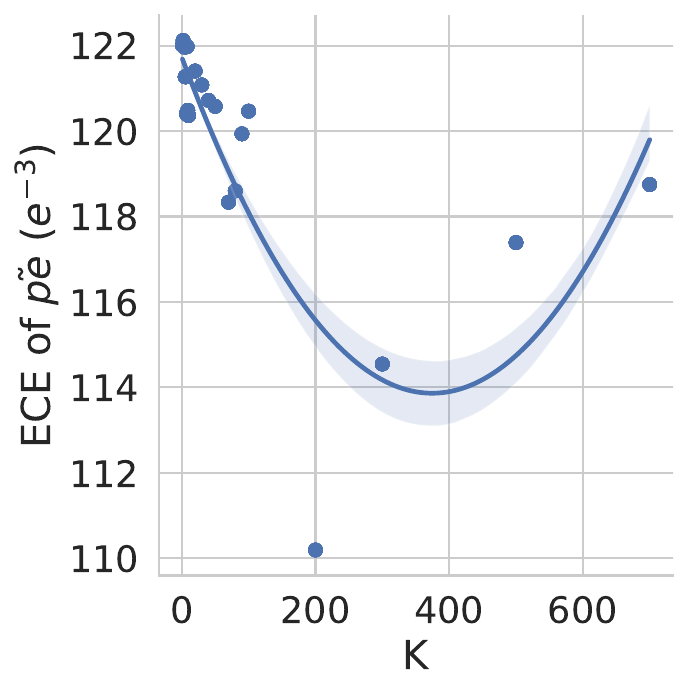}
    \caption{ECE of $\tilde{pe}$ generated from varying cluster numbers $K$ for the CIFAR100 dataset on T2T-ViT-14 with a fitted second order polynomial.}
    \label{fig:cifar100_ece}
\end{figure}

\paragraph{Clustering effectively separates predictions along entropy and probability of ground truth} We demonstrate that running K-means with large enough values of K effectively separates predictions across clusters corresponding to high/low entropies as well as clusters corresponding to high/low predicted probability of ground-truth. Figure \ref{fig:cluster_analysis_svhn_entropy_box_plot} shows entropy box-plots on SVHN using T2T-ViT-7 with K = 6 and K = 20. We can see that while the clusters with K = 6 have roughly the same average entropy, with K = 20 we start seeing more distinct clusters. A similar pattern can be seen when looking at the predicted probability of the ground-truth in figure \ref{fig:cluster_analysis_svhn_prob_gt_box_plot}. We show entropy per cluster on CIFAR100 on T2T-ViT-14 with K = 40 in figure \ref{fig:cifar100_k_40} and the ECE as a function of K in figure \ref{fig:cifar100_ece}.

\paragraph{ECE using $\widetilde{pe}$} In figure \ref{fig:cluster} we showed the expected calibration error computed on a separate test set when using $\widetilde{pe}$ as a function of K. $\widetilde{pe}$ corresponds to the ``delegate'' value of a a given cluster at a path $\pi$ as defined in \eqref{eq:approx}. 

To compute the ECE over the test set $\mathcal{S}_{\text{test}}$, we first compute the ECE at each path $\pi$ $\text{ECE}_{\pi}$ as follows:
\begin{enumerate}
    \item Obtain the prediction $\pred_{\pi}(x)$ for each sample $(x, y) \in \mathcal{S}_{\text{test}}$ at classifier $f_{\pi}$.
    \item Assign x to the appropriate cluster $Q^*$.
    \item Approximate the probability at $y$ as $\tilde{p}(y | x) = 1 - \widetilde{pe}(f_{\pi, Q^*}  |x_i)$.
    \item Following \citet{guo2017calibration}, we sort all $\tilde{p}(y | x)$ in ascending orders and split them into 15 bins denoted $B_1, B_2 \dots B_{15}$ 
    \item Compute the average predicted probability using $\tilde{p}(y | x)$ at each bin B as:
    \[\bar{P}(B) = \frac{1}{|B|} \sum_{i \in B} \tilde{p}(y_i | x_i)\]
    \item Compute the accuracy at each bin B: $ACC(B) = \frac{1}{|B|} \sum_{i \in B} \ind[\hat{y}_i = y_i ]$
    \item Compute the calibration error at each bin B  $\Delta(B)$ as the difference between the average predicted probability and the accuracy: $\Delta(B) = |ACC(B) - \bar{P}(B)|$
    \item The ECE for path $\pi$ is then calculated as $\text{ECE}_{\pi} = \sum_{b = 1}^{15} \frac{|Q_b|}{\mathcal{S}_{\text{test}}} \Delta(B_b)$
\end{enumerate}

We repeat the above procedure for each path $\pi \in \mathcal{P}$. The total ECE for a given $K$ is then computed as the average of $\text{ECE}_{\pi}$ for all paths:
\[
\text{ECE} = \frac{1}{|\mathcal{P}|}\sum_{\pi \in \mathcal{P}}\text{ECE}_{\pi}
\]

\subsection{Only predict the best step}\label{app:next_best}
Instead of predicting the $pe(f_{\pi'}|x)$ of each possible future paths $\pi'$ and take one step towards the best path, we can instead only predict one value per step. This would greatly reduce the complexity at inference from $O(B^EE)$ to $O(B)$. 
To do so, we replace our target $\tilde{pe}(f_{\pi'}|x) $  \textbf{to be the best possible value of the whole loss for a given step $b$,} which is defined as follows:
\begin{align}
    \ell^*_b = \argmin_{\pi' \in  \mathcal{P}_{\pi  \rightarrow b }} {\color{red} c_{\pi'}} + \tilde{pe}(f_{\pi'}|x)  \text{ the best loss value we can get for step }b.
\end{align}
This requires us to search amongst all future paths $\mathcal{P}_{\pi  \rightarrow b }$ and train our gate to directly predict that loss value  $\widehat{\ell}^*_b  = h^j_{\theta}(\info_j(x), b)$. This moves the $O(B^EE)$ inference time complexity to train time. At inference, the module predicts one value per step and the the decision is taken accordingly:
\begin{align}
     g^j_{\theta}() = \argmin_{b} \widehat{\ell}^*_b  .
\end{align}
Moreover, to train this model, we have to commit to a fixed cost/01 tradeoff ratio as it is integrated into the training through ${\color{red} c_{\pi'}}$. This means that we have to retrain a module for each acc/cost point.
\textbf{RMSE of the next best step approach}
In the ``next best step'' we directly predict loss value which contains a combination of the cost and the probability of error instead of only predicting probability of error.

This changes the domain of our target:
\begin{align}
 \text{previous target: }&\tilde{pe}(f_{\pi'}|x)  \in [0,1] \\
 \text{new target: }&\ell^*_b = \argmin_{\pi' \in  \mathcal{P}_{\pi  \rightarrow b }}  \lambda c_{\pi'} + \tilde{pe}(f_{\pi'}|x)  \in[-\lambda \min(c),\lambda \max(c)].
\end{align}
For $h$ ,we are using the same prediction function as before (see Eqn.~\ref{eqn:arch_h}), but without the sigmoid layer as the target is not a probability anymore.
\begin{figure}[h]
    \centering
    \label{fig:next_best}
   \includegraphics[scale=0.4]{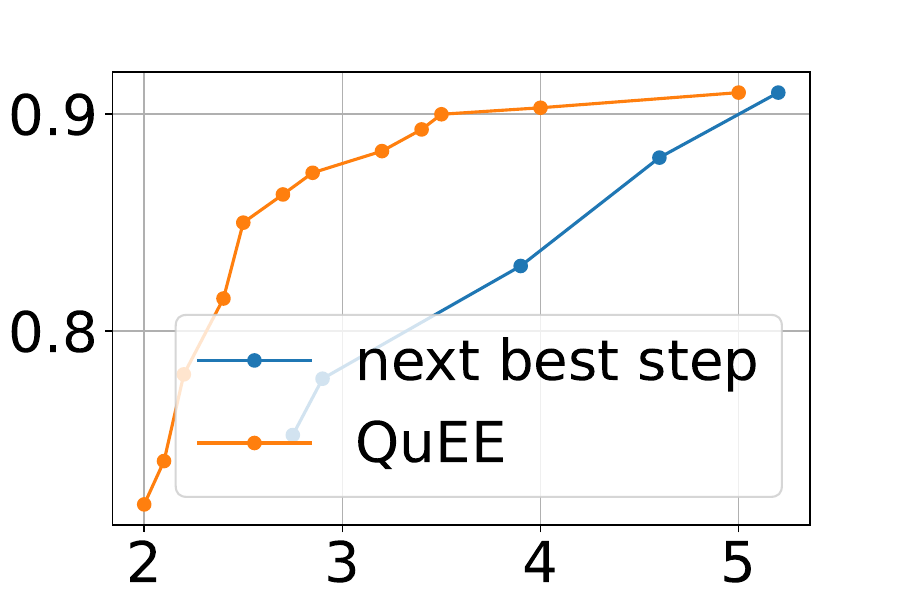}
\end{figure}
The results for this approach on one dataset are presented in Figure~\ref{fig:next_best}. This proposal is able to reach similar performance at a higher cost, but is quickly outperformed. In the very high-cost regime, the corresponding $\lambda$ values are set to a very low value in order to give all the importance to the performance term in the loss. This means that in this setting, the next best method is almost predicting the next best $\tilde{pe}(f_{\pi'}|x)$:
\begin{align}
\ell^*_b =& \argmin_{\pi' \in  \mathcal{P}_{\pi  \rightarrow b }}  \lambda c_{\pi'} + pe(f_{\pi'}|x)\\
\ell^*_b \approx& \argmin_{\pi' \in  \mathcal{P}_{\pi  \rightarrow b }}   pe(f_{\pi'}|x) \text{ if  } \lambda \text{ is small}.
\end{align}
However, the performance quickly degrades as we raise the importance of the cost through $\lambda$. This could indicate that the learning task of predicting the value of the next best step is considerably harder and that simple NN is unable to solve it.

\subsection{Expanded Literature Review}\label{app:exp_lit_review}

\paragraph{Dynamic neural networks} Dynamic architectures adapt their computational graphs to the input being processed \cite{han2022_dynamic_survey, compression_survey}. By adapting their depth (number of layers executed) or width (number of channels or neurons executed) for each sample, dynamic architectures can significantly reduce computation during inference \cite{han2022_dynamic_survey}.\\
Early-exit (EE) networks are a widely studied class of dynamic depth architecture where a prediction is obtained at an intermediate layer and subsquent layers are skipped \cite{bolukbasi_adaptive_nn, msdnet, dynamic_perc, regol2024_jeidnn, Ilhan2023AdaptiveDN}.  This is done by augmenting the network with intermediate inference modules at various layers. The idea was first presented in \citet{bolukbasi_adaptive_nn} where the augmented network is trained end-to-end. \citet{msdnet} addresses common issues arising from end-to-end training such as the interference of early inference modules on the performance of later ones by introducing architectural changes such as dense connections. \citet{dynamic_perc} further decouples feature extraction from classification by augmenting the network with an entire parallel stream for early classification while using the original backbone network for feature extraction. The two streams interact with each other via various attention mechanisms \cite{dynamic_perc}. However, end-to-end training is inconvenient when working with large foundation models \cite{efficient_ptq_lm, regol2024_jeidnn} and using a simple threshold is susceptible to miscalibration of earlier inference modules \cite{regol2024_jeidnn}. Post-training EE approaches instead rely on a fixed pre-trained backbone, and explore effective ways to train the inference modules and design more sophisticated gating mechanisms for the exit rule~\citep{regol2024_jeidnn,epnet_rl_learnable, Ilhan2023AdaptiveDN}. One of the challenges of early exiting is that the gating mechanism influences the samples reaching each inference modules. Failure to account for this effect can lead to a distribution shift at inference. This particular issue has first been studied by \citet{l2w, boostedNet} using a threshold-based exit mechanism. \citet{regol2024_jeidnn} propose a fixed-backbone EE procedure that uses a trainable gating mechanism leading to performance gains. However, one important drawback of \citet{regol2024_jeidnn} is that the training procedure needs to be repeated for every operating point making it impractical for use-cases where the computation budget changes over time.

Width-wise sample-adaptation can be achieved by selecting a subset of channels in CNN architectures as in \citet{huang2018condensenet, herrmann2020_channel_selection_gumbel}. A more general approach that is compatible with transformer-based architectures is SuperNets \cite{liu2018_d2nn, odena2017_changing_model_behavior_rl, hazimeh2020_tree_ensemble_layer} where a sample is dynamically routed through a subset of neurons at inference. \citet{hazimeh2020_tree_ensemble_layer} insert a differentiable decision tree layer in any neural network to benefit from the inherent conditional computation of decision trees while still being able to train the network end-to-end with backpropagation. \citet{odena2017_changing_model_behavior_rl, liu2018_d2nn} use reinforcement learning and gradient-based optimization to jointly-train the network augmented with controller modules that select the computational route for a sample.  Closer to our algorithm, works such as \citet{dual_dynamic_inference, fully_dynamic_inference_dnn} perform both depth and width adaptation via layer-skipping and channel-selection in convolution-based architectures. All these works rely on trainable controllers that are optimized jointly with the underlying network \cite{dual_dynamic_inference, fully_dynamic_inference_dnn} since the sub-networks need to be optimized for the inputs they handle. This makes them incompatible with large pre-trained foundation models. Foundation models are typically trained with very large proprietary datasets for extended periods of time \cite{oquab2024dinov2, gpt3, bommasani_foundation_models} in a self-supervised fashion. The idea of foundation models is to build a dataset-agnostic model that is trained once and can be readily reused on downstream datasets with only minor fine-tuning (typically of the inference head) \cite{oquab2024dinov2, bommasani_foundation_models}. For that reason, it is crucial to develop post-training efficiency techniques that do not require training of the backbone.

\paragraph{Quantization} Quantization is an effective way of speeding up inference where weights, gradients and activations of a model are represented at lower bit resolutions \cite{compression_survey}. Quantization-aware training (QAT) techniques quantize the network during training \cite{quant_survey, quant_survey_gholami} while post-training quantization (PTQ) is performed on a trained model with only a small amount of data \cite{quant_survey, quant_survey_gholami,dettmers2022_llmint8}. This makes PTQ particularly appealing as a width-compression technique when working with foundation models \cite{efficient_ptq_lm, dettmers2022_llmint8}. However, unlike the previously discussed width-adaptive techniques, quantization is typically not input-adaptive \cite{quant_survey, Liu2022_dqnet, hong2022cadyq}. Once a network has been quantized, the same structure is used for both ``easy'' and ``hard'' samples, employing unnecessarily high precision for easier samples \cite{Liu2022_dqnet, hong2022cadyq, Tian2023_cabm}. \citet{hong2022cadyq, Tian2023_cabm} consider dynamic mixed-precision quantization for image super-resolution. In that setting, a low-resolution image is split into patches which are super-resolved individually using a neural network. \citet{hong2022cadyq} introduce a bit selector that uses metrics of quantization sensitivity for each patch to determine the optimal bit width per layer. \citet{Tian2023_cabm} replace the bit-selector at inference with a look-up table indexed by the edge score, arguing that the edge score is a reliable measure for patch complexity in super-resolution tasks. Closer to our specific setting, DQNet \cite{Liu2022_dqnet} explores dynamic mixed-precision quantization for image classification. In DQNet, the network is quantized at different resolutions and is then augmented with a small neural network called the \textit{bit controller} whose function is to determine the bit resolution for subsequent layer blocks. The bit controller is fed with the intermediate feature map of a CNN \cite{Liu2022_dqnet}. While \citet{Liu2022_dqnet, hong2022cadyq, Tian2023_cabm} all dynamically adapt the bit precision on a per-sample basis, the fact that they encode the budget in their loss formulation means that the algorithm needs to be retrained for every operating point. They also do not exploit depth-adaptation.

\paragraph{Quantization of early-exit networks} Other works have combined the adaptability of early-exit networks with the efficacy of quantization by quantizing early-exit networks \cite{saxena2023_mcqueen, escepe}. In \citet{escepe} a pre-trained early-exit network is first split into sections and each section is quantized separately using weight-clustering. The quantized network is then fully retrained using knowledge distillation to recover the intermediate classifier's performance. \citet{saxena2023_mcqueen} uses a QAT approach where the optimal per-layer weights and activations quantization parameters are learnt during training. While both of these works combine quantization with early-exiting they both propose QAT-like approaches and are thus unsuitable for foundation models. They are also not sample adaptive along their widths as a single static mixed-precision quantization is learnt for all samples \cite{escepe, saxena2023_mcqueen}.



\end{document}